\documentclass[journal]{IEEEtran}
\usepackage{cite}
\usepackage{xspace} 
\usepackage{amsmath,amssymb,amsfonts} 
\usepackage{amsthm} 
\usepackage{hyperref} 
\usepackage{graphics} 
\graphicspath{{figs/}}
\usepackage{subfig} 
\usepackage{url} 
\usepackage{comment} 
\usepackage{bm}
\usepackage{multicol}
\usepackage{siunitx}
\usepackage{booktabs}
\usepackage{multirow}
\usepackage{bbm}
\usepackage{mathtools} 
\usepackage{etoolbox}
\usepackage{color}
\usepackage{authblk}

\DeclareMathOperator*{\argmax}{arg\,max}

\newcommand{\blue}[1]{\textcolor{black}{#1}}
\newcommand{\maung}[1]{\textcolor{black}{#1}}
\newcommand{\kinoshita}[1]{\textcolor{black}{#1}}
\newcommand{\RB}[1]{\textcolor{black}{#1}}
\newcommand{\both}[1]{\textcolor{black}{#1}}

\begin{document}

\title{An Overview of Compressible and Learnable Image Transformation with Secret Key and Its Applications}

\author[1]{Hitoshi Kiya}
\author[1]{AprilPyone MaungMaung}
\author[1]{Yuma Kinoshita}
\author[2]{Shoko Imaizumi}
\author[1]{Sayaka Shiota}
\affil[1]{Tokyo Metropolitan University, 6-6 Hino Tokyo, Japan}
\affil[2]{Chiba University, Japan} 
\maketitle

\begin{abstract}
This article presents an overview of image transformation with a secret key and its applications. Image transformation with a secret key enables us not only to protect visual information on plain images but also to embed unique features controlled with a key into images. In addition, numerous encryption methods can generate encrypted images that are compressible and learnable for machine learning. Various applications of such transformation have been developed by using these properties. In this paper, we focus on a class of image transformation referred to as learnable image encryption, which is applicable to privacy-preserving machine learning and adversarially robust defense. Detailed descriptions of both transformation algorithms and performances are provided. Moreover, we discuss robustness against various attacks.
\end{abstract}


\IEEEpeerreviewmaketitle

\section{Introduction}
Distributed systems for information processing such as cloud computing and edge computing have been spreading in many fields. However, the processing can lead to serious problems for end users, such as the unauthorized use of services, data leaks, and privacy being compromised due to unreliable providers and accidents~\cite{huang2014apsipatrans,lagendijk2013ieeespmagazine,moo2013nsdi,kiya2012signal}. In contrast, the spread of deep neural networks (DNNs) has greatly contributed to solving complex tasks for many applications, such as for computer vision, biomedical systems, and information technology~\cite{lecun2015deep}. Machine learning (ML) utilizes a large amount of data, which include sensitive personal information, to extract representations of relevant features so that the performance is significantly improved~\cite{tishby2015deep,saxe2019information}. However, there are also security issues when using ML in distributed systems to train and test data, such as \blue{compromised} data privacy, data leakage, and unauthorized access. Therefore, privacy-preserving ML has become an urgent challenge. In addition, DNN models are deployed in security-critical applications such as autonomous vehicles, healthcare, and finance due to their remarkable performance. The DNN models used in such applications have to be robust against various attacks such as model inversion attacks, membership inference attacks, and adversarial attacks~\cite{fredrikson2015model, shokri2017membership, 2014-ICLR-Szegedy, siva2020ieeespw}.

Many studies on secure, efficient, and flexible communication/storage/computing have been reported~\cite{siva2020ieeespw, wenjun2003ieeetmlti,ito2008icip, ito2008eusipco, ito2009eurasip, zhenjun2015mtapp,chai2021combining, chengqing2017ieeemm,erkin2007eurasip, watanabe2004fast, ferdowsi2020privacy}. Full encryption with provable security [like RSA (Rivest-Shamir-Adleman) and AES (Advanced Encryption Standard)] is the most secure option for securing multimedia data~\cite{lagendijk2013ieeespmagazine, aono2016privacy, araki2016high, araki2017optimized, lu2016using, shokri2015privacy, phong2018ieeetifs,phuong2019privacy, gilad2016cryptonets, wang2018efficient}, but there is a trade-off between security and other requirements such as for a low processing demand, bitstream compliance, and signal processing in the encrypted domain. Several perceptual encryption schemes have been developed to balance these trade-offs.

Accordingly, we present an overview of image transformation with a secret key, referred to as perceptional image encryption, which has \blue{has given new solutions to} the above issues. One way of privacy-preserving computing is to use a perceptual image encryption method that aims to protect visual information on plain images. Compared with number theory-based encryption~\cite{lagendijk2013ieeespmagazine, aono2016privacy, araki2016high, araki2017optimized, lu2016using, shokri2015privacy, phong2018ieeetifs,phuong2019privacy, gilad2016cryptonets, wang2018efficient}, such as multi-party computation and homomorphic encryption, perceptual encryption methods have a number of advantages. The use of perceptual encryption allows us to directly apply machine learning algorithms without increasing computational costs. In other words, there is no need to prepare algorithms specialized for computing encrypted data. 

Moreover, some perceptual encryption methods can produce compressible encrypted images, called compression-then-encryption (EtC) images~\cite{kenta2015pcs, Kenta2015ieice, Kenta2017ieice, chuman2019ieeetifs, zhou2014ieeetifs, Nimbokar2014ijca, liu2018ieeebio, liu2010ieeetip, hu2014icassp, johnson2004ieeetsp, gaata2016ijmter, osamu2015icassp}, and the use of perceptual encryption enables us to embed unique futures controlled with a key into an image~\cite{2020-Arxiv-Maung, maung2021protection,maung2020icip}. Accordingly, encrypted images can be designed so that they are compressible, learnable, or have a unique feature in addition to protecting visual information by using a perceptual encryption method. Thus, in this paper, we focus on a class of image transformation, referred to as learnable image encryption, that is applicable to privacy-preserving machine learning and adversarially robust defense.

\section{Image transformation with key and its applications}
\label{seq2}
\begin{figure*}[t]
 \begin{center}
 \includegraphics[scale=0.65]{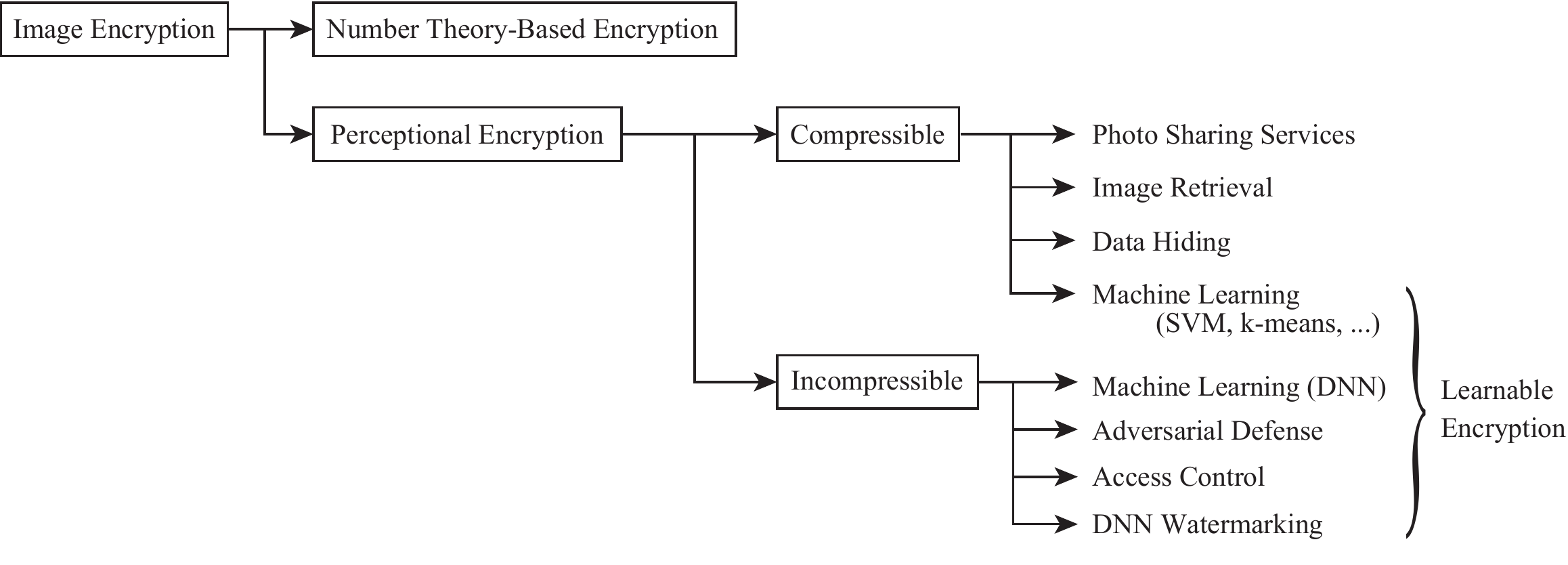}
 \caption{Perceptual image encryption methods and their applications.} 
 \label{fig:category}
 \end{center}
\end{figure*}
Figure~\ref{fig:category} shows a categorization of image transformation with a secret key, called image encryption or image cryptography, where the image transformation is classified into two classes: number theory-based encryption and perceptual encryption. Number theory-based encryption includes full encryption with provable security (like RSA and AES). In contrast, perceptional encryption can offer encrypted images that are described as bitmap images, so the encrypted images can be directly applied to image processing algorithms. In addition, encrypted images can be decrypted even when noise is added to them, although number theory-based \blue{encrypted images cannot}.

\begin{figure}[t]
\centering\includegraphics[width=\linewidth]{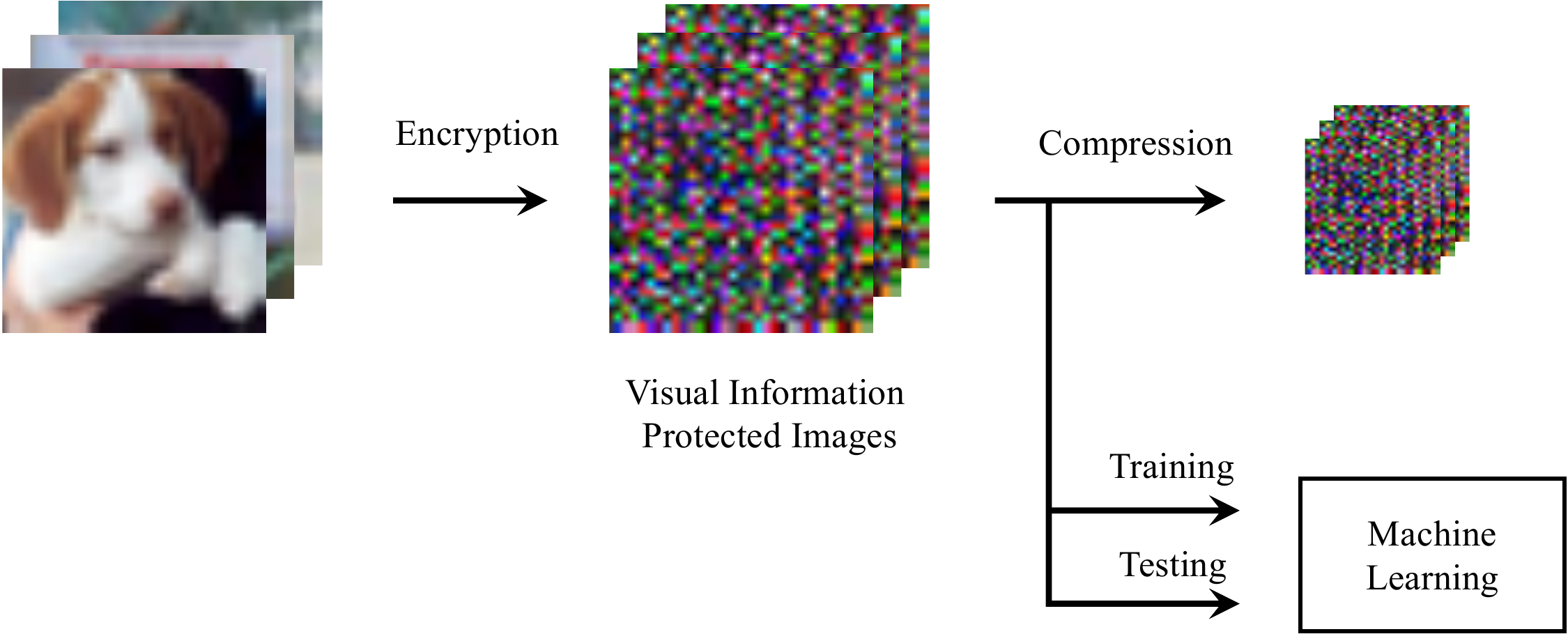}
\caption{Compressible and learnable image encryption.\label{fig:comp_and_lean}}
\end{figure}

In this paper, we focus on two properties of perceptually encrypted images: compressibility and learnability. As shown in Fig.~\ref{fig:comp_and_lean}, if an encrypted image can be compressed by using a compression method such JPEG compression, the encrypted image is compressible. \blue{In addition}, if an encrypted image can be applied to a learning algorithm such as DNNs in the encrypted domain, it is learnable. Image encryption prior to image compression is required in certain practical scenarios such as secure image transmission through an untrusted channel provider. An encryption-then-compression (EtC) system is used in such scenarios, although the traditional way of securely transmitting images is to use a compression-then-encryption (CtE) system. Compressible encryption is a key technology for implementing EtC systems.

Learnable encryption enables us to directly apply encrypted data to a model as training and testing data. Encrypted images have no visual information on plain images in general, so privacy-preserving learning can be carried out by using visually protected images. In addition, the use of a secret key allows us to embed unique features controlled with the key into images. From these properties, several transformation methods with a key have been proposed for adversarially robust defense, access control, and model watermarking.

\RB{In Section~\ref{seq3}, methods for generating compressible encrypted images are briefly summarized, and then the compressible encrypted images are also demonstrated to be learnable in Section~\ref{seq4}. Accordingly, the encrypted images can be applied to traditional ML such as support vector machine. Moreover, existing image encryption methods for privacy-preserving DNN models are compared in terms of image classification accuracy and robustness against various attacks in Sections~\ref{seq5}. In addition to applications to privacy-preserving processing, image transformation with a key is described to be applicable to robust defense against adversarial examples, and model protection from unauthorized access in Sections~\ref{sec:defense} and~\ref{seq7}, respectively. In these applications, image transformation aims to embed a unique feature into a model with a key, even though it aims to protect visual information on plain images for preserving privacy. Concluding remarks, the limitation and future work are in Section~\ref{sec:conclusion}.
}

\section{Compressible Image Encryption for EtC systems}
\label{seq3}
The origin of image transformation with a key is in block-wise image encryption schemes for EtC systems. Such compressible encryption methods are summarized here.

\subsection{Encryption-then-compression systems}
\begin{figure}[t]
\centering\includegraphics[width=\linewidth]{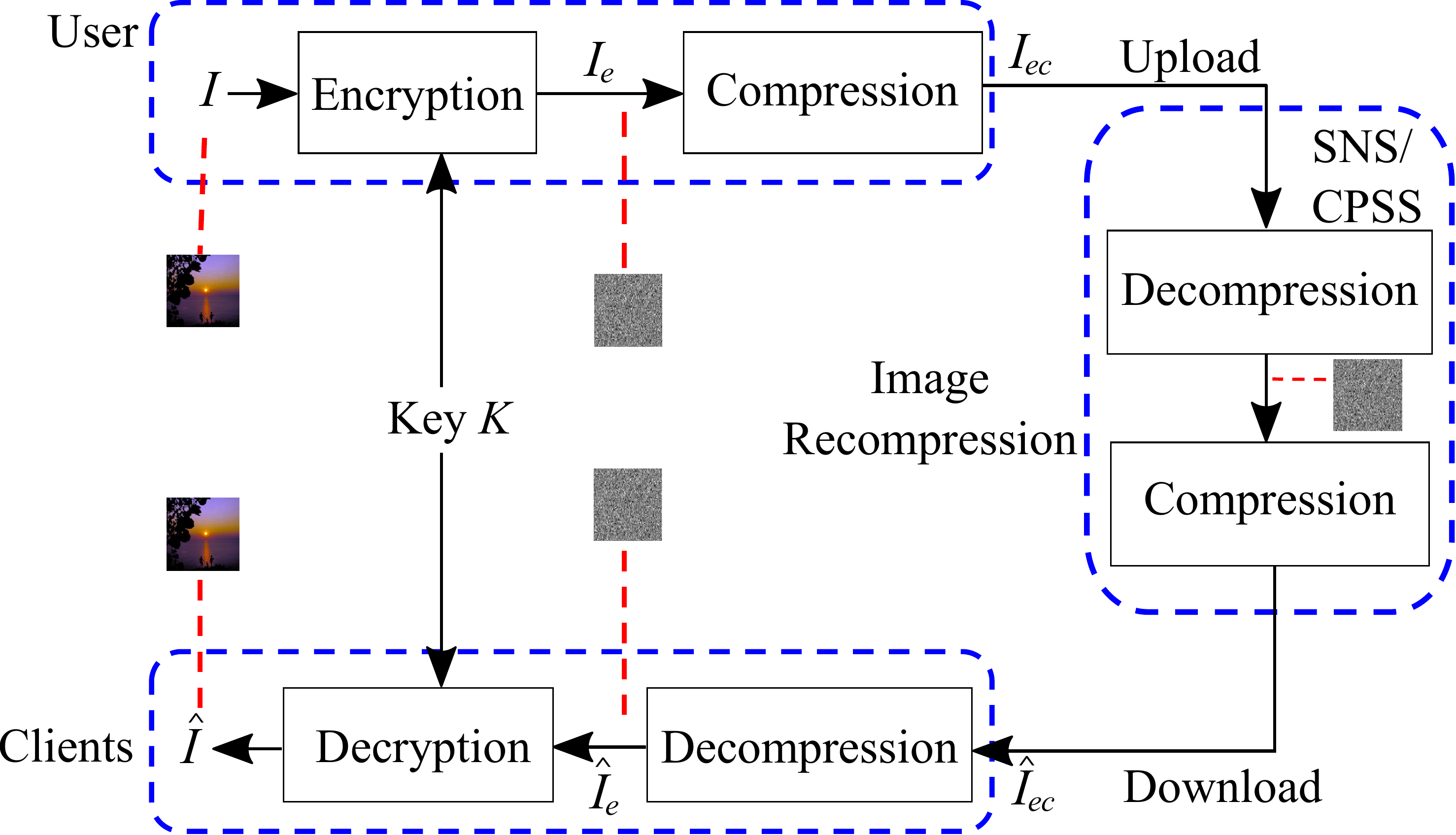}
\caption{EtC system.\label{fig:etcsystem}}
\end{figure}

Block-wise image encryption schemes~\cite{kenta2015pcs,Kenta2015ieice, Kenta2017ieice, chuman2019ieeetifs} have been proposed for EtC systems, in which a user wants to securely transmit an image $I$ to an audience or a client via social network services (SNS) or cloud photo storage services (CPSS), as shown in Fig.~\ref{fig:etcsystem}. The privacy of an image to be shared can be controlled by the user unless the user does not give the secret key $K$ to the providers, even when the image is generally recompressed by the providers. In contrast, in CtE systems, the disclosure of non-encrypted images or the use of plain images is required to recompress uploaded images. Accordingly, encrypted images to be applied to EtC systems have to be compressed multiple times by compression methods used by SNS providers such as JPEG \blue{compression} or other well-known compression methods used for SNS and CPSS. 

\begin{figure}[t]
\centering\includegraphics[width=\linewidth]{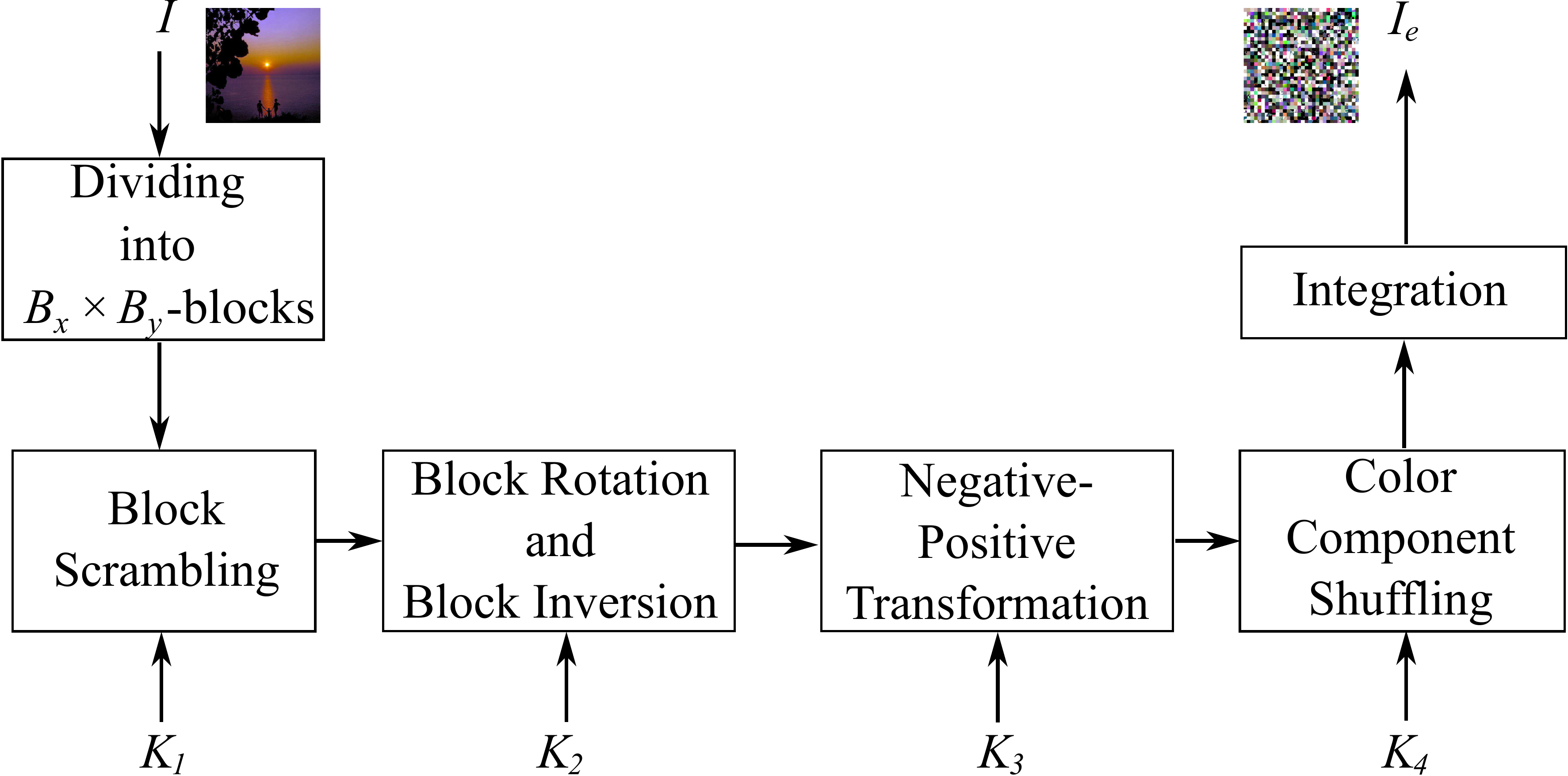}
\caption{Color-based block scrambling image encryption.\label{fig:color_bs}}
\end{figure}

\begin{figure}[t]
\captionsetup[subfigure]{justification=centering}
\centering
\subfloat[Block rotation]{
\includegraphics[clip, height=3.5cm]{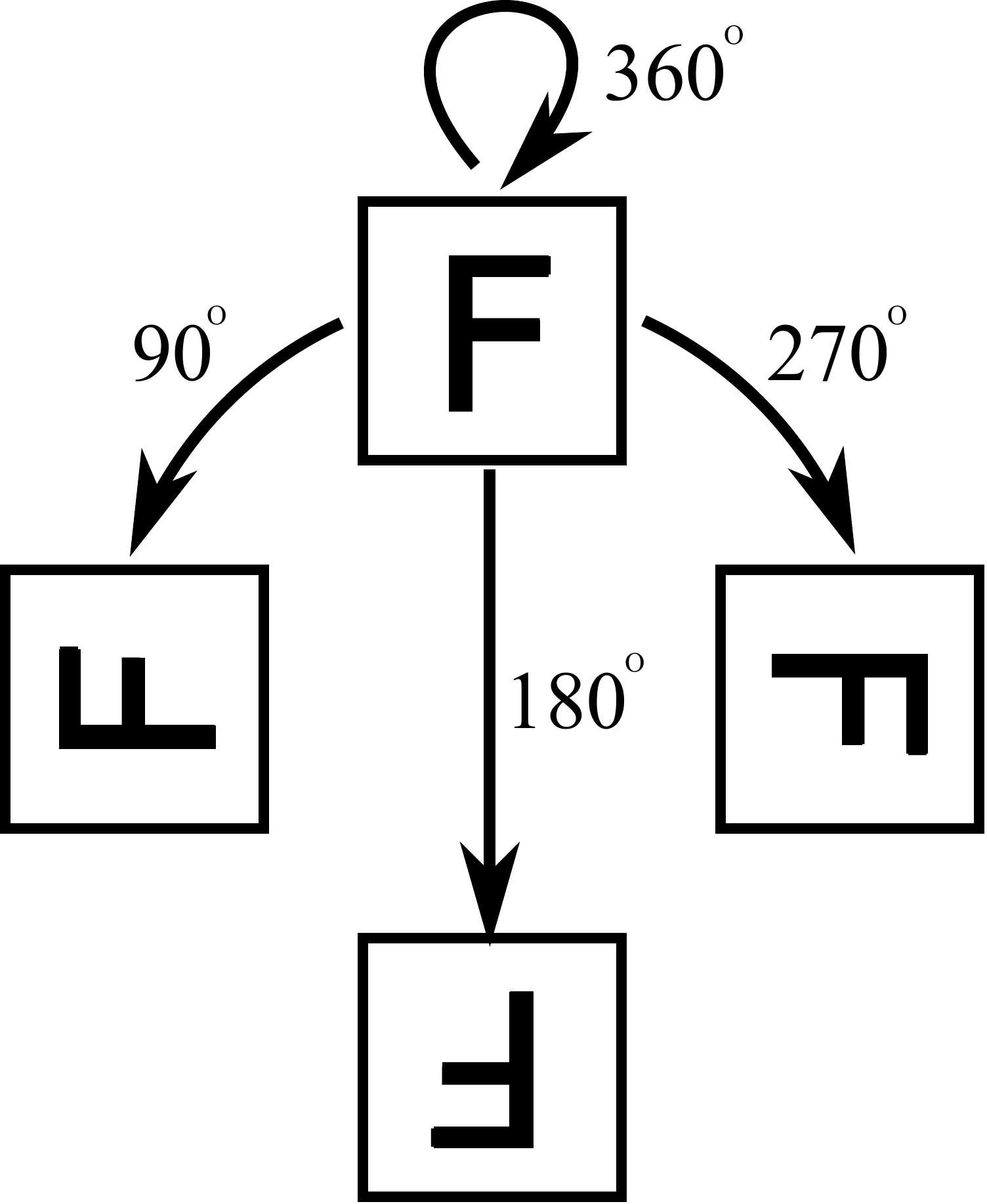}
\label{fig:rot}
}
\hspace{8mm}
\subfloat[Block inversion]{
\includegraphics[clip, height=3cm]{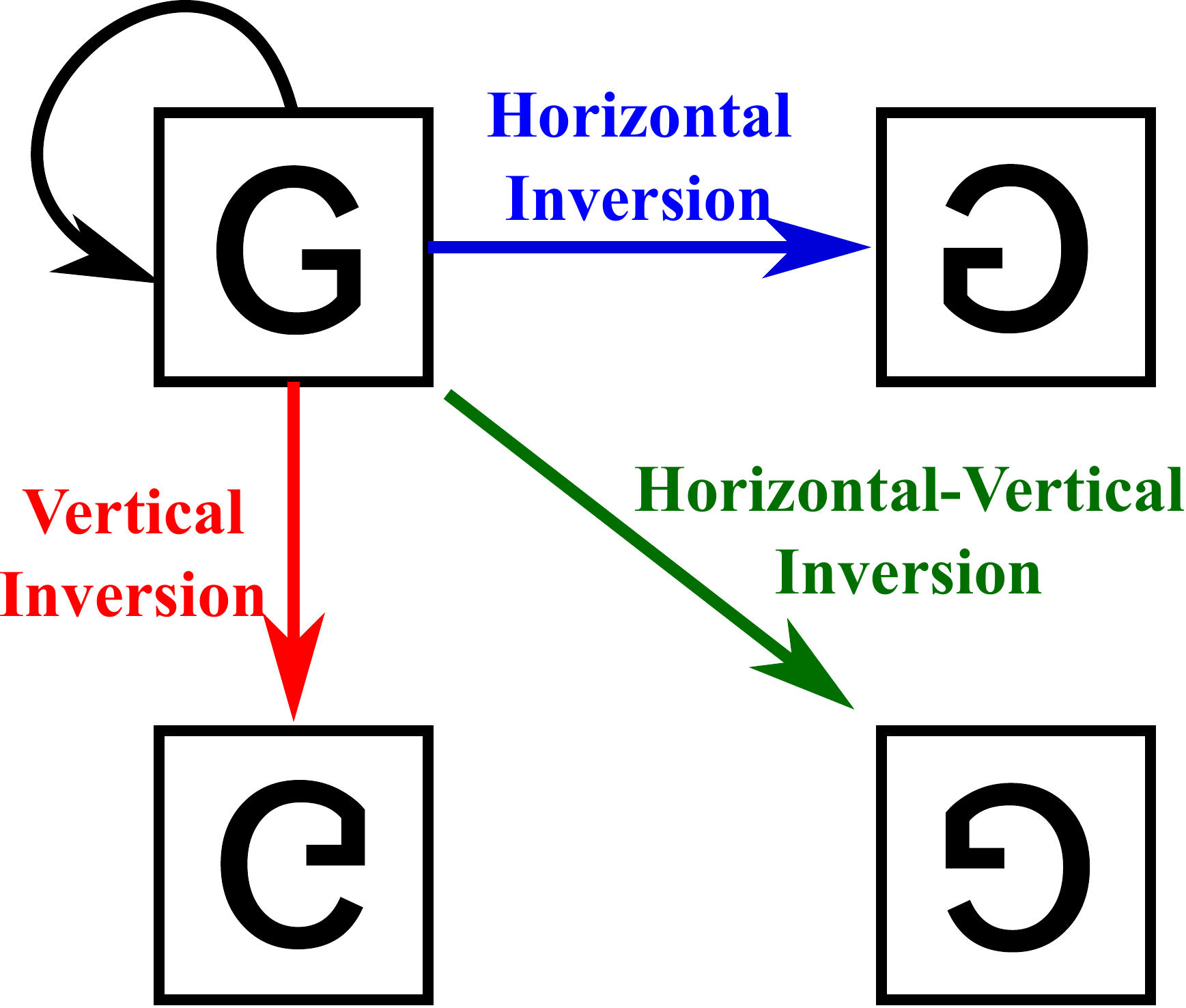}
\label{fig:inv}
}
\caption{Block rotation and inversion.}
\label{fig:rot_inv}
\end{figure}

\begin{figure}[t]
\centering
\includegraphics[width =8 cm]{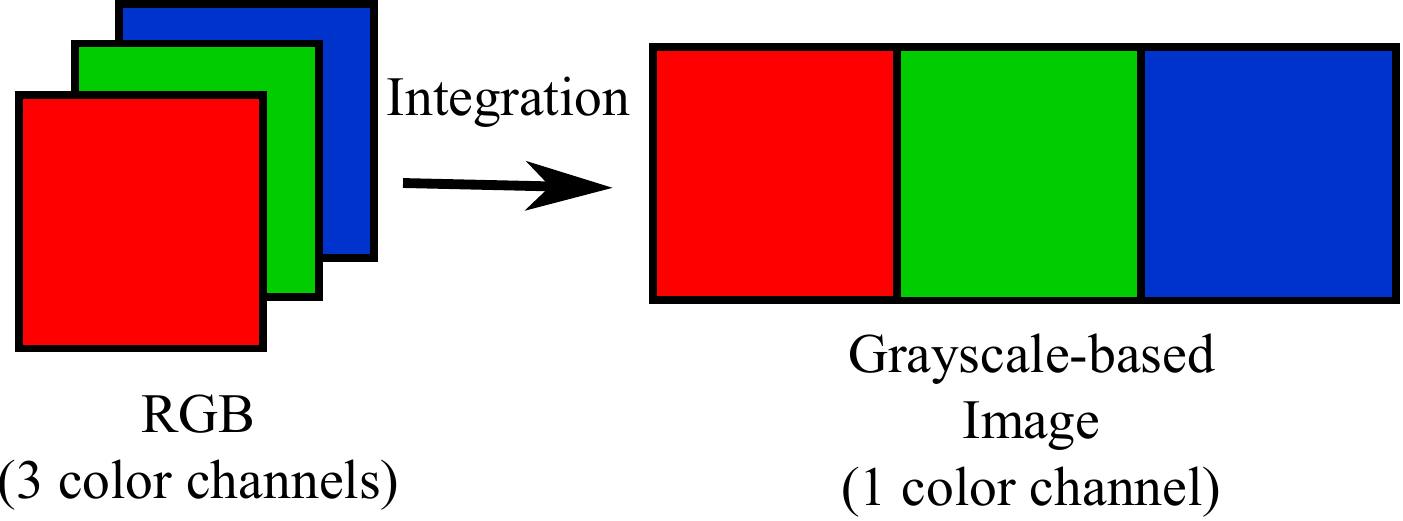}
\caption{Grayscale-based image generation in RGB space.}
\label{fig:gen_rgb}
\end{figure}

\begin{figure}[t]
\captionsetup[subfigure]{justification=centering}
\centering
\subfloat[$I_g$ without sub-sampling (4:4:4)]{\includegraphics[clip, width=8cm]{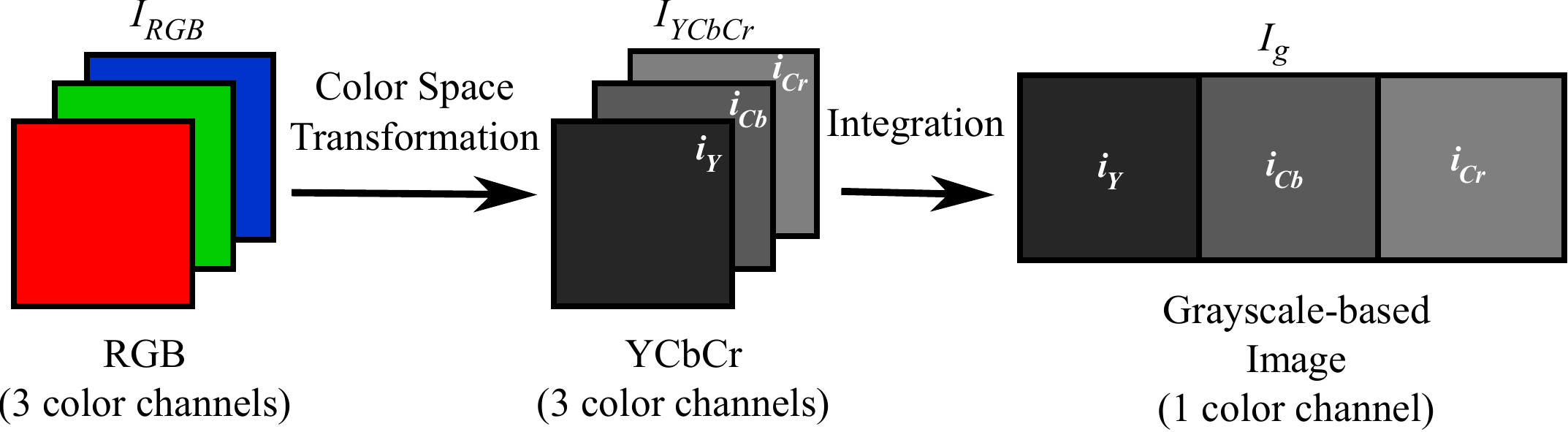}\label{fig:gen444}}
\\
\subfloat[$I_g$ with 4:2:0 sub-sampling]{\includegraphics[clip,
width=8cm]{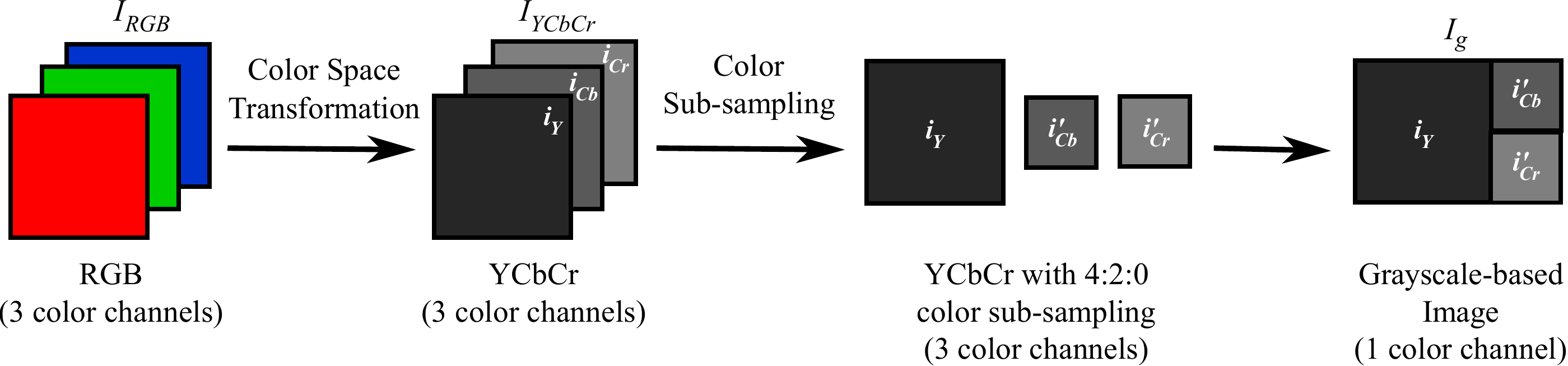}\label{fig:gen420}}
\caption{Grayscale-based image generation in YCbCr space.}
\label{fig:pro_gen}
\end{figure}

\begin{figure*}[t]
\captionsetup[subfigure]{justification=centering}
\centering
\subfloat[Original image ($X \times Y$ = $512 \times 512$)]{
\includegraphics[clip, height=2.7cm]{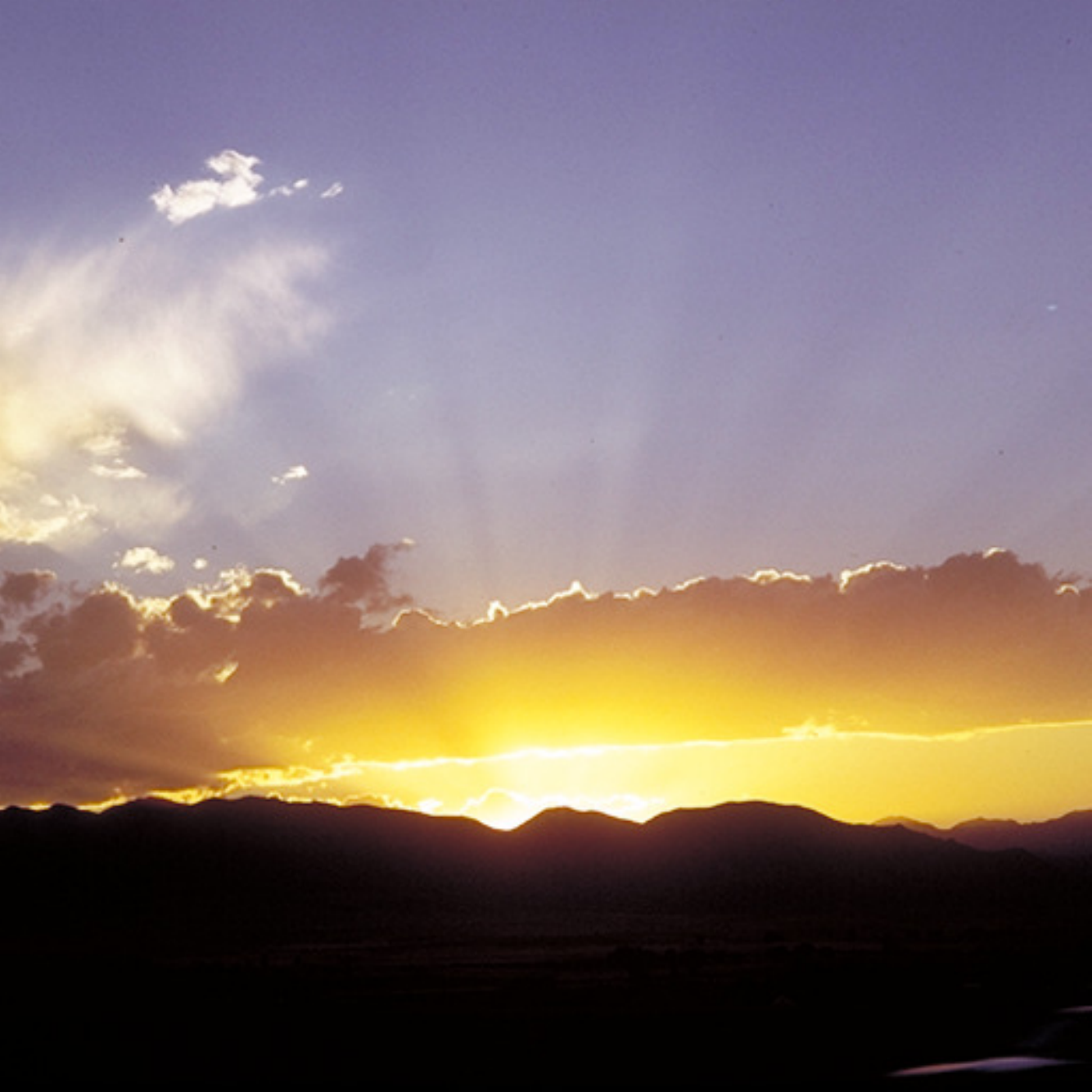}
\label{fig:ori}
}
\hspace{8mm}
\subfloat[Color-based scheme ($B_{x}=B_{y}=16$)]{
\includegraphics[clip, height=2.7cm]{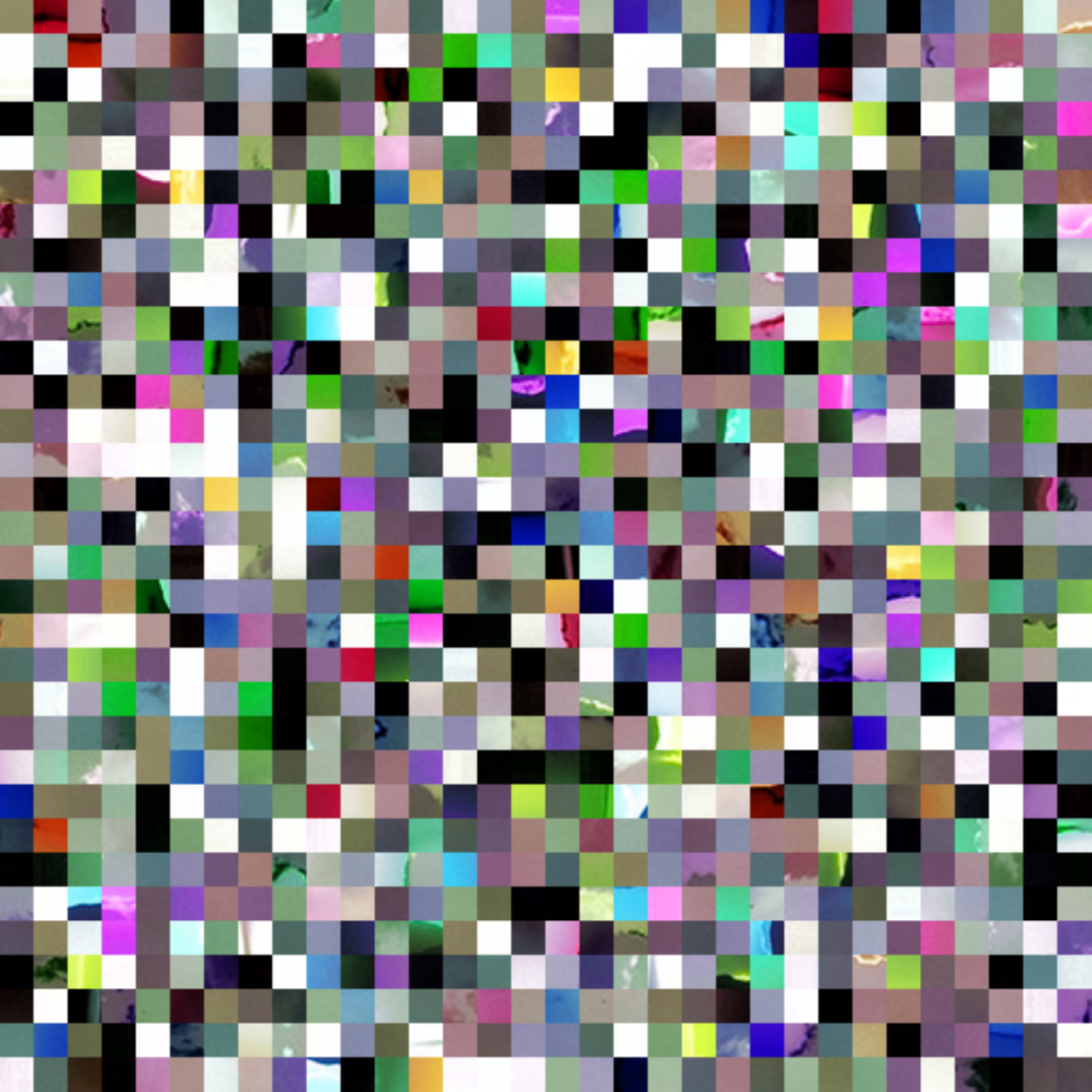}
\label{fig:col}
}
\hspace{8mm}
\subfloat[Grayscale-based scheme in RGB space ($B_{x}=B_{y}=8$)]{
\includegraphics[height=2.7cm]{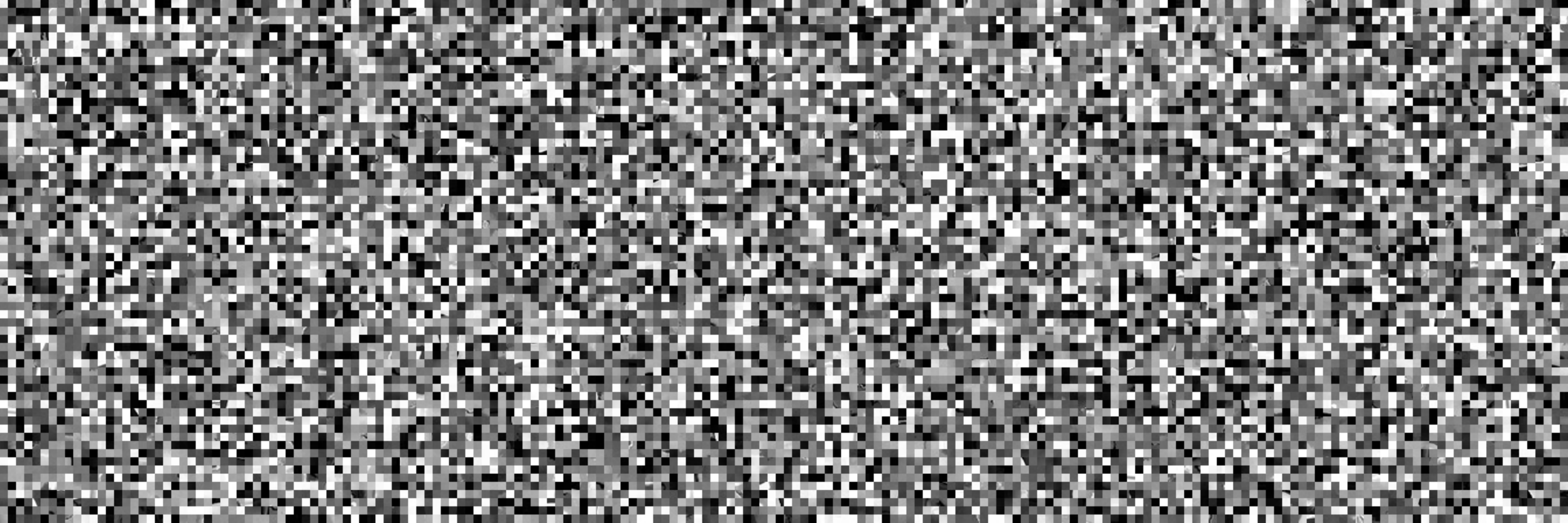}
\label{fig:conv}
}

\caption{Examples of images encrypted by using two encryption methods.}
\label{fig:examples_etc}
\end{figure*}

\subsection{Color-based image encryption}
The first \maung{encryption} method that can be used with JPEG compression is introduced here~\cite{kenta2015pcs}. This method was also demonstrated to be effective under the use of JPEG 2000~\cite{osamu2015icassp}, JPEG XR~\cite{kenta2016bmsb}, and lossless compression~\cite{Kenta2017ieice}.

A full-color image ($I$) with $X \times Y$ pixels is divided into non-overlapping blocks each with $B_x \times B_y$; then, four block scrambling-based encryption steps are applied to the divided blocks as follows (see Fig.~\ref{fig:color_bs}). 
\begin{description}
 \item[1) ] Randomly permute the divided blocks by using a random integer generated by a secret key $K_1$. 
 \item[2) ] Rotate and invert each block randomly (see Fig.~\ref{fig:rot_inv}) by using a random integer generated by a key $K_2$. 
 \item[3) ] Apply negative-positive transformation to each block by using a random binary integer generated by a key $K_3$, where $K_3$ is commonly used for all color components. In this step, a transformed pixel value in the $i$-th block $B_i$, $p'$, is calculated using 
\begin{align}
 p'=\begin{cases}
 {p \ \ \ \ \ \ \ \ \ \ \ \ \ \ \ \ \ \ \ \ \ \ (r(i) = 0)}\\
 {p\oplus(2^L-1) \ \ \ (r(i) = 1)}
 \end{cases},
 \label{eq:negaposi}
\end{align}
where $r(i)$ is a random binary integer generated by $K_3$, $p \in B_i$ is the pixel value of the original image with $L$ bits per pixel, and $\oplus$ is the bitwise exclusive-or operation. The value of occurrence probability $P(r(i)) = 0.5$ is used to invert bits randomly.
 \item[4) ] Shuffle three color components in each block by using an integer randomly selected from six integers generated by a key $K_4$ as shown in Table~\ref{tab:color_perm}.
\end{description}

\begin{table}
\caption{Permutation of color components for random integer. For example, if random integer is equal to 2, red component is replaced by green one, and green component is replaced by red one, while blue component is not replaced.\label{tab:color_perm}}
{\begin{tabular}{|>{\centering\arraybackslash}m{1in}||>{\centering\arraybackslash}m{1cm}|>{\centering\arraybackslash}m{1cm}|>{\centering\arraybackslash}m{1cm}|}
\hline
\multirow{2}{*}{Random Integer} & \multicolumn{3}{c|}{Three Color Channels
}\\
& R & G & B\\
\hline
0 & R & G & B\\
\hline
1 & R & B & G\\
\hline
2 & G & R & B\\
\hline
3 & G & B & R\\
\hline
4 & B & R & G\\
\hline
5 & B & G & R\\
\hline
\end{tabular}}{}
\end{table}

An example of an encrypted image with $B_x = B_y = 16$ is shown in Fig.~\ref{fig:examples_etc}(b), where Fig.~\ref{fig:examples_etc}(a) is the original one. Images encrypted by using color-based image encryption have almost the same compression performance as non-encrypted ones when using JPEG compression with $B_x = B_y = 16$~\cite{Kenta2015ieice,chuman2019ieeetifs}. Images encrypted by using block-wise encryption are called EtC images.

\subsection{Security against ciphertext-only attacks}
Security mostly refers to protection from adversarial forces. Most image transformation methods are designed to protect visual information that allows us to identify an individual, a time, and the location of a taken photograph. Untrusted providers and unauthorized users are assumed to be adversaries. Block-wise encryption has to be robust against both brute-force and jigsaw puzzle solver \blue{attacks used as} ciphertext-only attacks~\cite{chuman2017icme, chuman2018ieice}.

\begin{enumerate}
\item Brute-force attack\\
When an image with $X \times Y$pixels is divided into blocks with $B_x \times B_y$ pixels, the number of blocks n is given by
 \begin{align}
 n=\biggl\lfloor \dfrac{X}{B_x}\times \dfrac{Y}{B_y} \biggr\rfloor,
 \end{align}
where $\lfloor \cdot \rfloor$ is a function that rounds down to the nearest integer. The four block scrambling-based processing steps in Fig.~\ref{fig:color_bs} are then applied to the divided image. The key space of the color-based encryption $N_C(n)$ is given as below~\cite{Kenta2015ieice},
\begin{align}
 N_C(n)=N!\times8^n\times2^n\times6^n.
\end{align}
For example, when a color image with $1024\times 768$ pixels is divided into $16\times 16$ blocks, we obtain $n=3072$, and $N_C(3072)=3072!\times8^{3072}\times2^{3072}\times6^{3072}$.

\item Jigsaw puzzle solver attack\\
The jigsaw puzzle solver is a method of assembling jigsaw puzzles by using a computer~\cite{son2014solving,paikin2015solving,son2016solving,sholomon2016automatic}. In block-wise encryption, if we regard the blocks as pieces of a jigsaw puzzle, decrypting encrypted images is similar to assembling a jigsaw puzzle. Therefore, jigsaw puzzle solvers should be considered as one of the attack methods against block-wise encryption. Extended jigsaw puzzle solvers for block-wise image encryption~\cite{chuman2017icme,chuman2018ieice} have been proposed to assemble encrypted images including rotated, inverted, negative-positive transformed, and color component shuffled blocks. It has been shown that assembling encrypted images becomes difficult when the encrypted images are under the following conditions.
\begin{itemize}
\item Number of blocks is large.
\item Block size is small.
\item Encrypted images include compression distortion.
\item Encrypted images have less color information.
\end{itemize}
Since jigsaw puzzle solvers utilize color information to assemble puzzles, reducing the number of color channels in each pixel makes assembling encrypted images much more difficult. Thus, grayscale-based encryption schemes, which will be described next, have a higher security level than that of the color-based scheme because the number of blocks is large, the block size is small, and there is less color information.
\end{enumerate}

Other attacking strategies such as the known-plaintext attack (KPA) and chosen-plaintext attack (CPA) should be considered for security. Block-wise encryption becomes robust against KPA through the assigning of a different key to each image for encryption. In addition, the keys used for encryption do not need to be disclosed because the encryption scheme is not public key cryptography. Therefore, the encryption can avoid the CPA, unlike public key cryptography.

\subsection{Grayscale-based image encryption}
Two grayscale-based image encryption schemes~\cite{chuman2019ieeetifs, sirichotedumrong2019grayscale} were proposed to enable the use of a smaller block size and a larger number of blocks, which enhances both invisibility and security against several attacks. Furthermore, images encrypted by using the grayscale-based schemes include less color information due to the use of grayscale images, which makes the EtC system more robust. Figure~\ref{fig:gen_rgb} shows a grayscale-based image generated with the first grayscale-based image encryption scheme~\cite{chuman2019ieeetifs}, where a color image is split into three (RGB) channels, and the three channels are then combined to generate one grayscale-based image. Figure~\ref{fig:pro_gen} also illustrates a grayscale-based image \blue{generated by} the second grayscale-based image encryption scheme~\cite{sirichotedumrong2019grayscale}, where RGB components are transformed into the YCbCr color space, and the three transformed channels are then combined to generate one grayscale-based image. After defining a grayscale-based image, three encryption steps from step 1) to 3) are applied to the grayscale-based image in a similar manner to the color-based one. 

In Fig.~\ref{fig:pro_gen}, the use of the YCbCr color space allows us not only to improve the JPEG compression efficiency but also to avoid the effect of sub-sampling for chroma components, even when JPEG images with a 4:2:0 sub-sampling ratio are interpolated to increase the spatial resolution for chroma components in the decoding process [see Fig.~\ref{fig:pro_gen}(b)]. SNS and CPSS providers are known to manipulate uploaded images by changing the sub-sampling ratio and JPEG quality factor Q~\cite{chuman2017image,chuman2019image}, so users cannot choose a desired sub-sampling ratio and a value for Q in general.

\subsection{Applications of EtC images}
EtC images generated by using an image transformation method with a secret key have several interesting properties such as being compressible, learnable, and visually protected, so various applications of EtC images have been developed as shown in Fig.~\ref{fig:category}. EtC images were initially proposed to be applied to SNS or CPSS for privacy protection. In addition, they were demonstrated to be applicable to privacy-preserving reversible data hiding, image identification, and image retrieval~\cite{imaizumi2020reversible,kenta2020ieeeaccess,iida2019image}. Moreover, since EtC images are learnable as described in Section~\ref{seq4}, a number of machine learning algorithms, such as support vector machine (SVM) with a kernel trick, k-nearest neighbor (kNN), and random forests, can be carried out directly by using encrypted images~\cite{ayana2020ieice}. Besides, space modeling and dictionary learning with encrypted data can be performed~\cite{bandoh2020ieeeaccess, nakachi2020secure, nakachi2020privacy}. 

\section{Learnable image transformation for traditional machine learning}
\label{seq4}
\begin{figure}
\centering
\includegraphics[width=\linewidth]{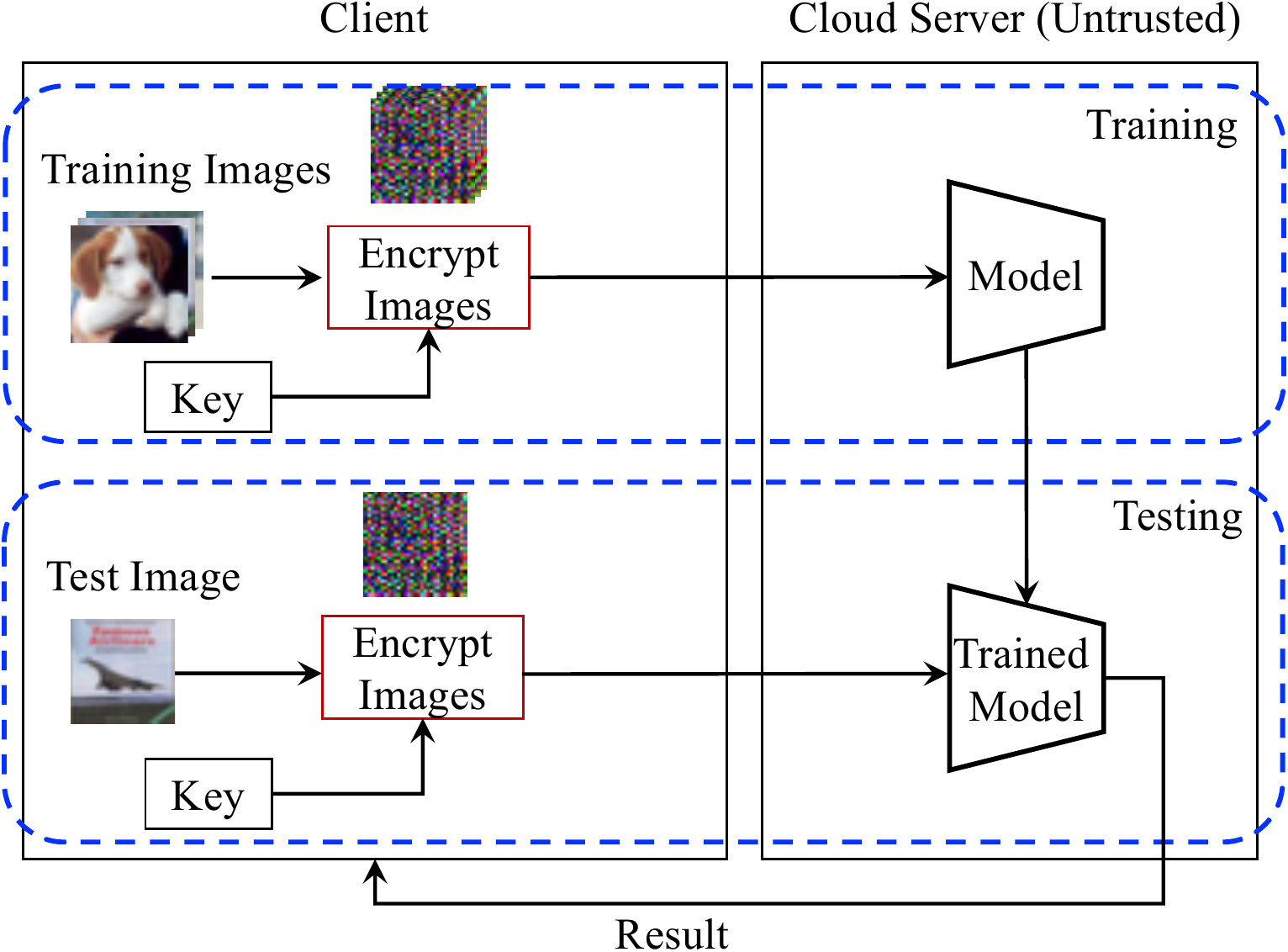} 
\caption{Privacy-preserving machine learning.\label{fig:ppml}}
\end{figure}

In this section, we discuss the application of EtC images to traditional machine learning algorithms such as support vector machines (SVM), k-nearest neighbor ($k$-NN), and random forests for privacy-preserving machine learning~\cite{ayana2020ieice}. As shown in Fig.~\ref{fig:ppml}, a model is trained by using training data (images) encrypted with a common key, and test images encrypted with the key are then applied to the trained model. Properties of EtC images are shown here, and the properties enable us to carry out privacy-preserving machine learning without any performance degradation~\cite{ayana2020ieice}.

\RB{
In Fig.~\ref{fig:ppml}, the goal of privacy-preserving machine learning is to classify encrypted images without any visual information in an untrusted cloud server. In this scenario, encrypted images have to be robust enough against various attacks.
}

\subsection{Traditional machine learning}
Traditional machine learning models are trained on the basis of the relationships between feature vectors of images, e.g., distance, inner product, and order relationship of elements. Here, we briefly show that SVM, $k$-NN, and random forests are based on such relationships.

\subsubsection{Linear SVM}
We first focus on linear SVM for two-class classification. In SVM computing, an input feature vector $\bm{x}$ is classified as
\begin{equation}
 \hat{y} = 
 \begin{cases}
 1 & (f(\bm{x}) > 0) \nonumber \\
 -1 & (f(\bm{x}) < 0) \nonumber
 \end{cases}.
\end{equation}
By using a weight vector $\bm{w}$ and bias $b$, the decision function $f(\bm{x})$ is given as
\begin{equation}
 f(x) = \bm{w}^\top \bm{x} + b.
 \label{svm}
\end{equation}
The training of SVM, i.e., obtaining decision function $f$ from a given dataset $S = \{(\bm{x}_i, y_i) | 1 \le i \le N\}$, is done by solving the following dual problem with respect to a dual variable vector $\bm{\alpha} = (\alpha_1, \cdots, \alpha_N)$,
\begin{align}
 & \max_{\bm{\alpha}} \: - \frac{1}{2} \sum_{1 \le i, j \le N} \alpha_i \alpha_j y_i y_j \bm{x}_i^\top \bm{x}_j + \sum_{1 \le i \le N} \alpha_i \nonumber \\
 & \mathrm{s.t.} \sum_{1 \le i \le N} \alpha_i y_i = 0, \:
 0 \le \alpha_i \le C,
 \label{eq:svm_problem}
\end{align}
where $C$ is a regularization parameter for the margin, and $y_i$ is a true label ($1$ or $-1$) of $\bm{x}_i$. The weight vector $\bm{w}$ and the bias $b$ are calculated by using the optimum $\bm{\alpha}$ as
\begin{align}
 \bm{w} &= \sum_{1 \le i \le N} \alpha_i y_i \bm{x}_i \\
 b &= y_{i'} - \sum_{1 \le i \le N} \alpha_i y_i \bm{x}_i^\top \bm{x}_i, \nonumber \\
 & i' \in \{i | 1 \le i \le N \land 0 < \alpha_i < C\}.
\end{align}

As shown above, Eq. (\ref{eq:svm_problem}) depends on the inner product of the input feature vectors, not feature vectors themselves.

\subsubsection{SVM with kernel trick}
SVM also has a well-known technique called the ``kernel trick'' for non-linear classification. For non-linear classification using SVM, a function $\phi$ that maps an input vector $\bm{x}$ onto high dimensional feature space $\mathcal{F}$ is utilized.

When the function $\phi$ is applied, the decision function in Eq. (\ref{svm}) becomes
\begin{equation}
 f(x)={\rm sign}(\bm{w}^\top \phi(\bm{x})+b), 
\end{equation}
and the optimization problem in Eq. (\ref{eq:svm_problem}) is given by
\begin{align}
 & \max_{\bm{\alpha}} \: - \frac{1}{2} \sum_{1 \le i, j \le N} \alpha_i \alpha_j y_i y_j \phi(\bm{x}_i)^\top \phi(\bm{x}_j) + \sum_{1 \le i \le N} \alpha_i \nonumber \\
 & \mathrm{s.t.} \sum_{1 \le i \le N} \alpha_i y_i = 0, \:
 0 \le \alpha_i \le C.
 \label{eq:kernel_svm}
\end{align}
In the kernel trick, the kernel function $K(\bm{x}_i,\bm{x}_j)=\phi(\bm{x}_i)^\top \phi(\bm{x}_j)$ of two vectors $\bm{x}_i$, $\bm{x}_j$ is defined instead of directly defining function $\phi$.

Typical kernel functions are the radial basis function (RBF) kernel and the polynomial one. RBF kernel is based on the Euclidean distance, and the polynomial kernel is based on inner products:
\begin{equation}
 K(\bm{x}_i,\bm{x}_j)={\rm exp}(-\gamma\|\bm{x}_i-\bm{x}_j\|^2)
 \label{rbfkernel}
\end{equation}
\begin{equation}
 K(\bm{x}_i,\bm{x}_j)=(1+\bm{x}_i^\top \bm{x}_j)^l,
\end{equation}
where $\gamma$ is a hyperparameter for deciding the complexity of boundary determination, and $l$ is a parameter for deciding the degree of the polynomial.

\subsubsection{$k$-NN}
The $k$-NN algorithm is usually based on the Euclidean distance. For each sample $\bm{x}_i$ in a training dataset $S$, the Euclidean distance
\begin{equation}
 \|\bm{x}_i-\bm{x}\|^2
\end{equation}
between testing samples $\bm{x}$ and $\bm{x}_i$ is first calculated. Then, $k$ nearest neighbors are picked up in accordance with the calculated Euclidean distance. Testing sample $\bm{x}$ is classified in the class most common among the neighbors. The predicted label $\hat{y}$ will be
\begin{equation}
 \hat{y} = \argmax_{q\in\{1, -1\}} k_q,
\end{equation}
where $k_q$ indicates the number of samples whose class label is $q$ among the $k$ nearest neighbors.

\subsubsection{Decision tree and random forests}
\label{subsec:rf}
Decision trees and random forests~\cite{breiman2001random} are learned on the basis of the order relationship of features among samples in a training dataset. In other words, the separation boundary of decision trees does not depend on the scale and bias of the features. Specifically, any full-rank diagonal matrix $\bm{D} \in \mathbb{R}^{d \times d}$ and any vector $\bm{b} \in \mathbb{R}^d$ can be used to transform the training dataset $S=\{(\bm{x}_i, y_i)\}$ as $S'=\{(\bm{D} \bm{x}_i + \bm{b}), y_i\}$, and a decision tree trained with $S'$ provides exactly the same results as that trained with $S$.

\subsection{Properties of EtC images}
Let us transform the $i$-th grayscale image $I_i$ with $H \times W$ pixels into a vector $\bm{x}_i=(p(0, 0), \cdots, p(H-1, W-1))^\top \in \mathbb{R}^d, d=H \times W$, where $p(h, w), 0 \le h \le H-1, 0 \le w \le W-1$ is a pixel value at the $(h, w)$ of $I$.

\begin{figure}
 \centering
 \includegraphics[width=0.95\linewidth]{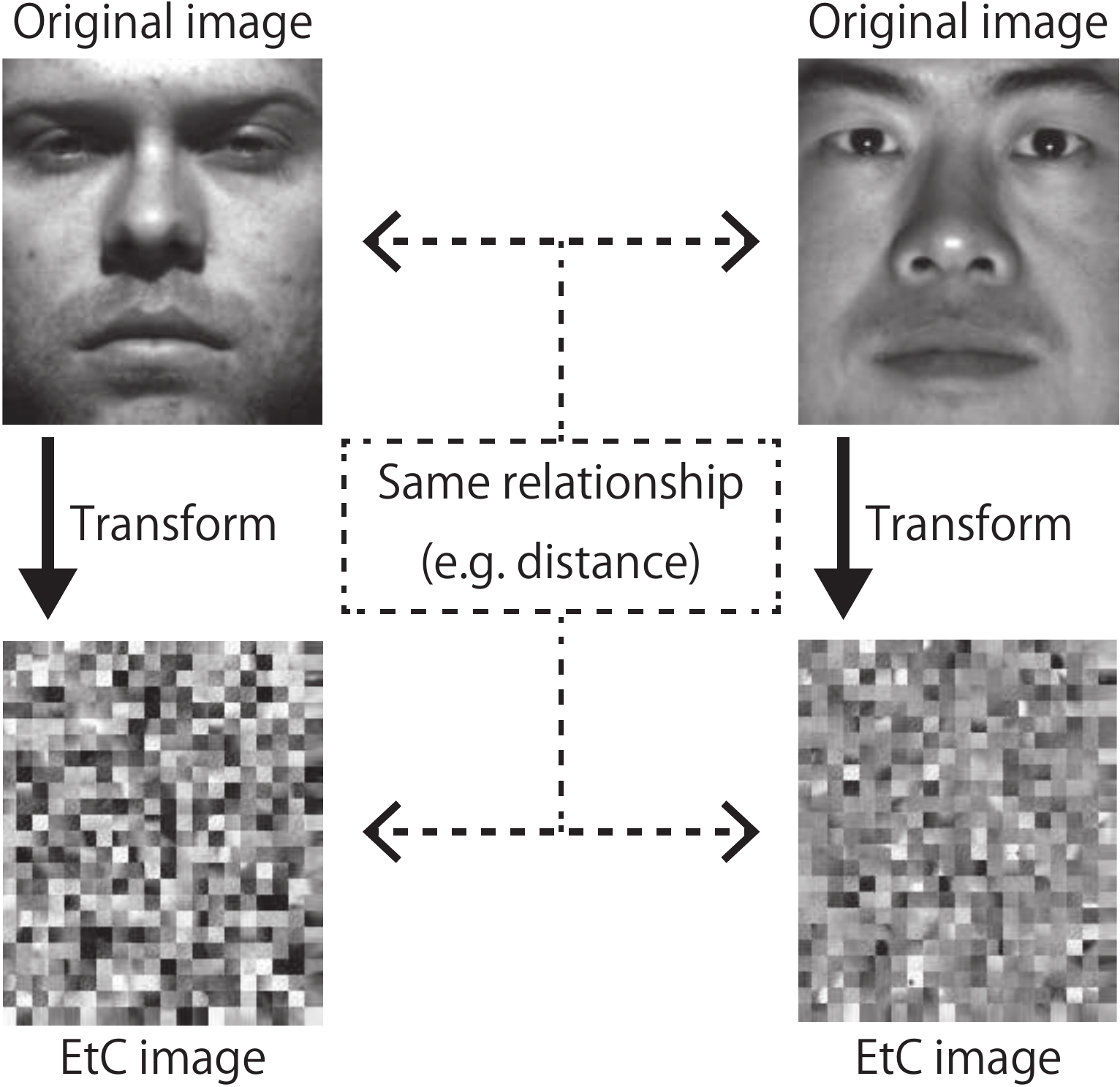}
 \caption{Relationship between two EtC images. Even after image transformation, typical relationship between two EtC images is same as that between corresponding original images.}
 \label{fig:etc_properties}
\end{figure}

EtC images have three properties, that is, the Euclidean distance, the inner product, and the order relationship of features between original vectors $\bm{x}_i$ and $\bm{x}_j$ are preserved after the image transformation (see Fig. \ref{fig:etc_properties}). Because the linear SVM, SVM with the kernel trick, $k$-NN, and random forests are based on the Euclidean distance, the inner product, and the order relationship of features, respectively, the properties of EtC images enable us to carry out privacy-preserving machine learning without any performance degradation~\cite{ayana2020ieice}.

\subsubsection{Block scrambling, block rotation and inversion}
As shown in Fig.~\ref{fig:color_bs}, block scrambling and block rotation and inversion are carried out to permute pixels. These operations are easily shown to be represented as a permutation matrix. For example, a permutation matrix $\bm{Q}$ is given as, for $d=3$,
\begin{equation}
 \bm{Q} = \left(
 \begin{array}{ccc}
 1 & 0 & 0 \\
 0 & 0 & 1 \\
 0 & 1 & 0 \\
 \end{array}
 \right),
\end{equation}
where $\bm{Q}$ has only one element of 1 in each row or each column, and the others are 0, so $\bm{Q}$ becomes an orthogonal matrix. The orthogonal matrix meets the equation
\begin{equation}
 \bm{Q}^\top \bm{Q}=\bm{E},
 \label{Q}
\end{equation}
where $\bm{E}$ is an identity matrix.

A encrypted vector $\hat{\bm{x}}_i$ is computed by using $\bm{Q}$ as
\begin{equation}
 \hat{\bm{x}}_i=\bm{Q}\bm{x}_i.
 \label{QBI}
\end{equation}
Therefore, $\hat{\bm{x}}_i$ in Eq. (\ref{QBI}) meets the properties in Eqs. (\ref{kyori}) and (\ref{naiseki}) due to the orthogonality of $\bm{Q}$~\cite{nakamura2016unitary,maekawa2019privacy}, where $\hat{p}_{i,j}(k)$ corresponds to a pixel value of an EtC image generated under the use of block scrambling, and block rotation and inversion operations.

\noindent Property 1: Conservation of Euclidean distances:
\begin{equation}
 \|\bm{x}_i-\bm{x}_j\|^2=\|\hat{\bm{x}}_i-\hat{\bm{x}}_j\|^2.
 \label{kyori}
\end{equation}

\noindent Property 2: Conservation of inner products:
\begin{equation}
 \bm{x}_i^\top \bm{x}_j = \hat{\bm{x}}_i^\top \hat{\bm{x}}_j.
 \label{naiseki}
\end{equation}
$\bm{x}_j$ is a transformed vector from the $j$-th image $I_j$.

\subsubsection{Negative-positive transformation}
Next, let us consider the influence of negative-positive transformation. In the case of using the transformation in Eq. (\ref{eq:negaposi}), the relation between a pixel value $p'_i(h, w) = p_i(h, w)$ and another $p'_j(h, w) = 255 - p_j(h, w)$ is given by
\begin{equation}
 (p'_i(h, w) - p'_j(h, w))^2 = (p_i(h, w) - p_j(h, w))^2.
\end{equation}
Note that $p_i(h, w)\oplus (2^L-1)$ is equal to $255 -p_i(h,w)$ when $L=8$. From this relation, it is confirmed that the Euclidean distance between $\bm{x}_i$ and $\bm{x}_j$ is preserved after negative-positive transformation. However, since the relation
\begin{align}
 p'_i(h, w) \cdot p'_j(h, w) &= (255 - p_i(h, w)) \cdot (255-p_j(h, w)) \nonumber \\
 &\neq p_i(h, w) \cdot p_j(h, w),
\end{align}
the inner product $\bm{x}_i^\top \bm{x}_j$is not preserved. Consequently, the negative-positive transformation operation preserves only the Euclidean distance between $\bm{x}_i$ and $\bm{x}_j$. In addition, negative-positive transformation can be written in the form $\bm{D} \bm{x}_i + \bm{b}$ using a full-rank diagonal matrix $\bm{D}$ and a vector $\bm{b}$, so it does not affect the order relationship of the features as shown in Section \ref{seq4}. \ref{subsec:rf}).

\subsubsection{Negative-positive transformation with z-score normalization}
We can preserve the inner product even under the use of negative-positive transformation by using z-score normalization~\cite{jain2005score}, which is a well-known data normalization method for machine learning. In z-score normalization, a value $p_i(h, w)$ is replaced with $z_i$ like
\begin{equation}
 z_i(h, w) = (p_i(h, w) - \overline{p}(h, w)) / \sigma(h, w),
\label{normalization}
\end{equation}
where $\overline{p}(h, w) = \frac{1}{N} \sum_{i=1}^N p_i(h, w)$, and $\sigma(h, w)$ is a standard deviation given by
\begin{equation}
 \sigma(h, w) = \sqrt{\frac{1}{N} \sum_{i=1}^{N}(p_i(h, w) -\overline{p}(h, w))^2}.
\end{equation}
Therefore, in negative-positive transformation, Eq. (\ref{normalization}) is given as
\begin{align}
 z'_i(h, w) &= -\frac{p'_i(h, w) - \overline{p'}(h, w)}{\sigma'(h, w)} \nonumber \\
 &= \frac{(255 - p_i(h, w)) - (255 - \overline{p}(h, w)}{\sigma'(h, w)} \nonumber \\
 &= -\frac{p_i(h, w) - \overline{p}(h, w)}{\sigma(h, w)} = -z_i(h, w),
 \label{znp}
\end{align}
where
\begin{equation}
 \overline{p'}(h, w) = \frac{1}{N} \sum_{i=1}^{N} p'_i(h, w) = 255 - \overline{p}(h, w),
\end{equation}
and
\begin{align}
 \sigma'(h, w) &= \sqrt{\frac{1}{N} \sum_{i=1}^{N}(p'_i(h, w) - \overline{p'}(h, w))^2} \nonumber \\
 &= \sqrt{\frac{1}{N} \sum_{i=1}^{N} (-p_i(h, w) + \overline{p}(h, w))^2} = \sigma(h, w).
\end{align}
Eq. (\ref{znp}) means that the normalized value $z'_i(h, w)$ of $p'_i(h, w)$ becomes the sign inverted value of the normalized value $z_i(h, w)$ of $p_i(h, w)$. A sign inversion matrix can be expressed as an orthogonal matrix, so both the Euclidean distance and the inner product are preserved under the use of z-score normalization. In addition, z-score normalization does not affect the order relationship of the features.

Hence, in the case of applying z-score normalization to EtC images, negative-positive transformation allows us to maintain the inner products. As a result, EtC images can maintain the Euclidean distance, the inner product, and the order relationship of features under the use of z-score normalization.

\subsubsection{Experimental results}
A face authentication simulation was carried out. We used the Extended Yale Face Database B~\cite{georghiades2001few}, which consists of 38 $\times$ 64 = 2432 frontal facial images with 192 $\times$ 160 pixels for $N=38$ persons. $M=64$ images for each person were divided in half randomly for training data samples and queries. $B_x \times B_y=8\times 8$ was used.

\begin{figure}
 \centering
 \includegraphics[width=0.95\linewidth]{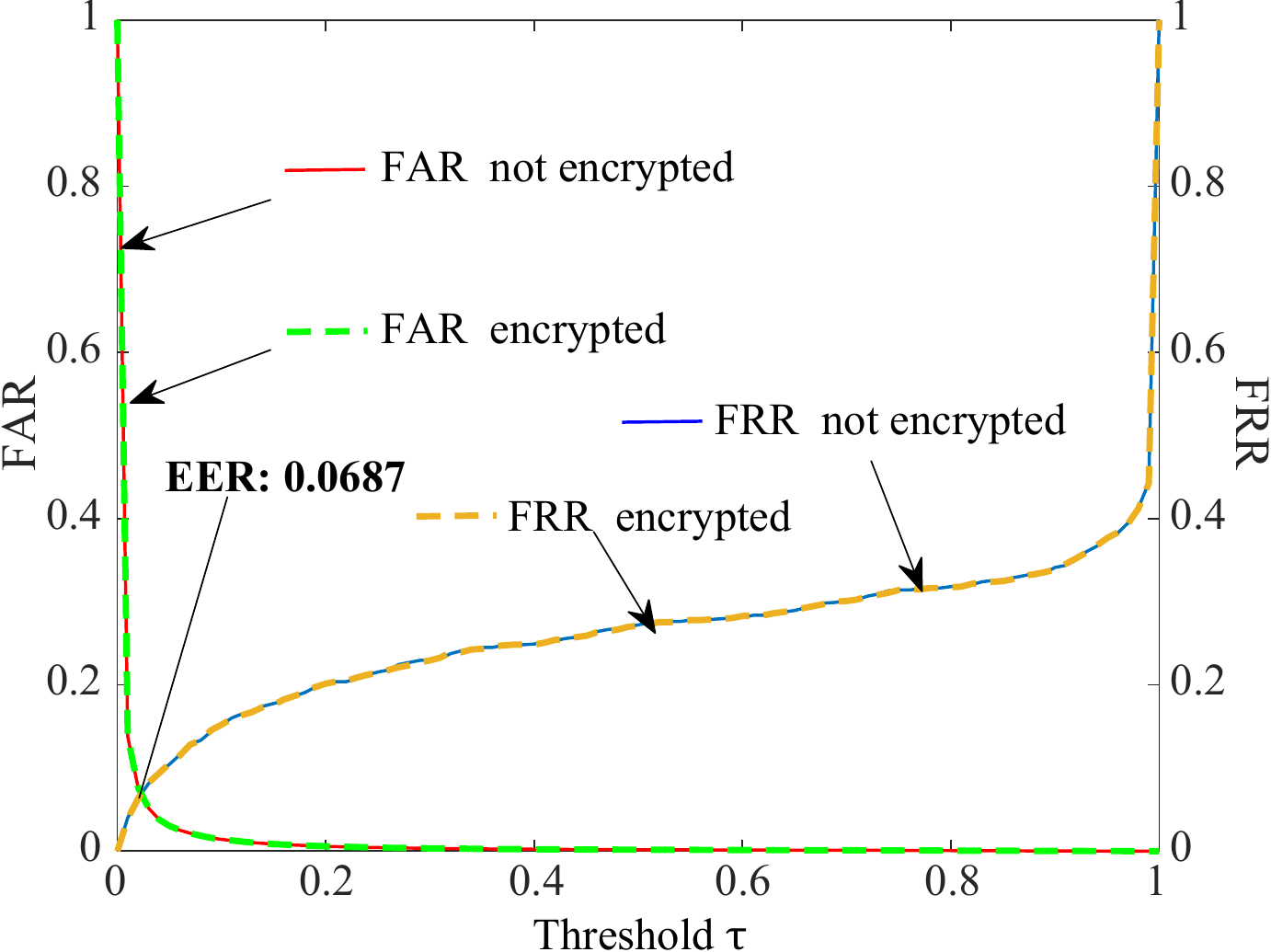}
 \caption{\kinoshita{Experimental results with SVM.}}
 \label{fig:my_label}
\end{figure}

In the simulation, we trained SVM classifiers using the RBF kernel in Eq. (\ref{rbfkernel}) with z-score normalization, and the false rejection ratio (FRR) and the false acceptance ratio (FAR) were calculated under a threshold $\tau$ for classification. From the results illustrated in Fig. \ref{fig:my_label}, the EtC images were confirmed to have no influence on the performance of the SVM classifiers under z-score normalization.

In other simulations, we also confirmed that dimensionality reduction methods, i.e., random projection~\cite{kaski1998dimensionality} and random block sampling, can be carried out in the encrypted domain to reduce the number of dimensions of feature vectors,
\both{as shown in our previous work~\cite{ayana2020ieice}}.
Therefore, dimensionality reduction methods can be applied to representations for privacy-preserving machine learning. 

\section{Learnable image transformation for DNN}
\label{seq5}
\begin{figure}
\centering
\subfloat[]{\includegraphics[clip, height=2.5cm]{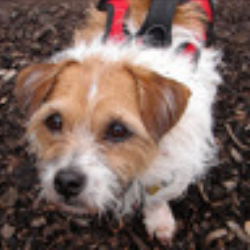}%
\label{fig:org}}
\hfil
\subfloat[]{\includegraphics[clip, height=2.5cm]{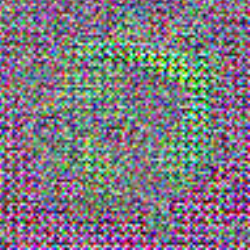}%
\label{fig:tanaka}}
\hfil
\subfloat[]{\includegraphics[clip, height=2.5cm]{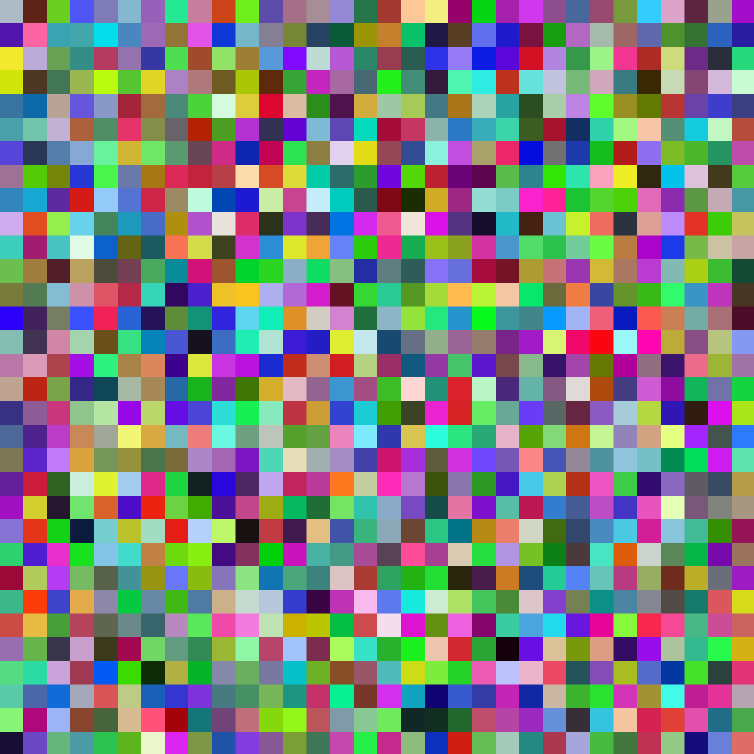}%
\label{fig:ele}}\\
\subfloat[]{\includegraphics[clip, height=2.5cm]{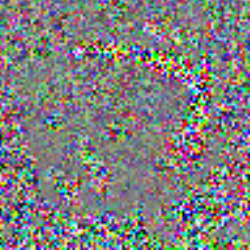}%
\label{fig:pixel}}
\hfil
\subfloat[]{\includegraphics[clip, height=2.5cm]{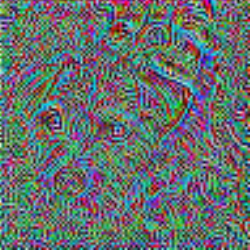}%
\label{fig:gan}}
\hfil
\subfloat[]{\includegraphics[clip, height=2.5cm]{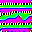}%
\label{fig:henkan}}
\caption{Example of images generated by various learnable image transformation methods for DNNs. (a) Original image. (b) Tanaka~\cite{2018-ICCETW-Tanaka}. (c) E-Tanaka~\cite{madono2020block}. (d) Pixel-based~\cite{2019-Access-Warit}. (e) GAN-based~\cite{sirichotedumrong2020gan}. (f) U-Net-based~\cite{ito2021image}.\label{fig:examples-lie}}
\end{figure}

In this section, we present learnable image transformation methods that are specifically designed for privacy-preserving DNNs since EtC images used for traditional machine learning cannot maintain the high performance that using plain images achieves. Learnable image transformation methods for DNNs are classified into three types: block-wise transformation, pixel-wise transformation, and network-based transformation. Figure~\ref{fig:examples-lie} shows an example of images transformed by various learnable transformation methods.

\subsection{Learnable image transformation}
Figure~\ref{fig:ppml} depicts a framework for learnable image transformation for DNNs, which is the same as that for traditional ML. Transformed images without visual information are sent to a cloud server for training and testing a model, and the network in the cloud server classifies the images without being aware of any visual information. Three types of image transformation are summarized below.

\subsubsection{Block-wise transformation}
Tanaka first introduced a learnable image transformation method that works in a block-wise manner for image classification as learnable image encryption~\cite{2018-ICCETW-Tanaka}, where a block-wise adaptation layer is used prior to the classifier to reduce the influence of image encryption. In Tanaka's method, a color image is divided into blocks, and each block is processed by using pixel shuffling (upper and lower 4-bit split pixels) and negative/positive transformation with a common key for all blocks. 

Next, to enhance the security of encryption, the method was extended by adding a block scrambling step as used for EtC images~\cite{madono2020block}. Hereinafter, we refer to this extended learnable transformation as ``E-Tanaka.'' The E-Tanaka method allows us to assign a different key to each block.

\subsubsection{Pixel-wise transformation}
A pixel-wise transformation method was proposed~\cite{2019-Access-Warit} in which negative-positive transformation and color component shuffling are applied. It enables us not only to carry out data augmentation in the encrypted domain but also to use independent keys for training a model and testing. In addition, this pixel-wise transformation does not need any adaptation layer prior to the classifier. 

\subsubsection{Network-based transformation}
Another type of learnable image transformation for DNNs is network-based transformation that uses generative models to generate visually protected images.
\kinoshita{In network-based methods, a generative model producing protected images is trained by considering both classification accuracy for a classifier and perceptual loss based on a VGG model. Therefore, the generative model is optimized to remove visual information on plain images while maintaining a high classification accuracy.}
One network-based method utilizes a generative adversarial network (GAN)~\cite{sirichotedumrong2020gan}. Images encrypted by using this method can be used for both training and test images. In contrast, a transformation method with U-Net~\cite{ito2021image} was proposed to enhance robustness against various attacks, but it cannot be used for training a model.
Figure~\ref{fig:examples-lie} shows an example of images transformed by various learnable transformation methods.

\subsection{Comparison among learnable encryption methods}
Learnable encryption for DNNs still has a number of issues that should be solved because of its short history. To clarify the issues, existing learnable encryption methods are compared in terms of design architecture, classification accuracy, robustness against attacks, and restriction of use.

\begin{figure*}[t!]
 \centering
 \begin{tabular}{ccccccc}
 Attack method & Tanaka &
 E-Tanaka &
 Pixel-based & GAN-based & U-Net-based \\
 &
 \cite{2018-ICCETW-Tanaka} &
 \cite{madono2020block} &
 \cite{2019-Access-Warit} &
 \cite{sirichotedumrong2020gan} &
 \cite{ito2021image} \\\\[-5pt]
 Protected & 
 \begin{minipage}{2.5cm}
 \centering
 \includegraphics[width=2.5cm]{figs/tanaka_363}
 \end{minipage} &
 \begin{minipage}{2.5cm}
 \centering
 \includegraphics[width=2.5cm]{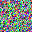}
 \end{minipage} &
 \begin{minipage}{2.5cm}
 \centering
 \includegraphics[width=2.5cm]{figs/pixel_363}
 \end{minipage} &
 \begin{minipage}{2.5cm}
 \centering
 \includegraphics[width=2.5cm]{figs/gan_363}
 \end{minipage} &
 \begin{minipage}{2.5cm}
 \centering
 \includegraphics[width=2.5cm]{figs/protected.png}
 \end{minipage}\\
 & 0.037 & 0.002 & 0.041 & 0.164 & 0.020 \\[5pt]
 FR-Attack \cite{fr-attack} & 
 \begin{minipage}{2.5cm}
 \centering
 \includegraphics[width=2.5cm]{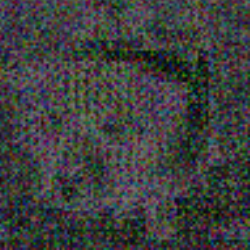}
 \end{minipage} &
 \begin{minipage}{2.5cm}
 \centering
 \includegraphics[width=2.5cm]{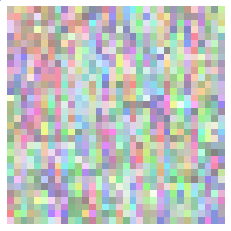}
 \end{minipage} &
 \begin{minipage}{2.5cm}
 \centering
 \includegraphics[width=2.5cm]{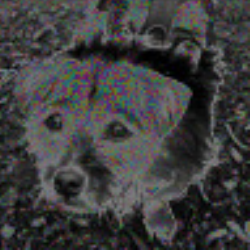}
 \end{minipage} &
 \begin{minipage}{2.5cm}
 \centering
 \includegraphics[width=2.5cm]{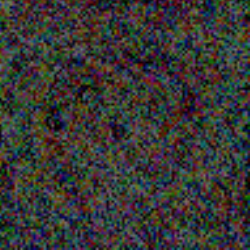}
 \end{minipage} &
 \begin{minipage}{2.5cm}
 \centering
 \includegraphics[width=2.5cm]{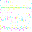}
 \end{minipage} \\
 & 0.101 & 0.055 & 0.303 & 0.091 & 0.007 \\[5pt]
 GAN-Attack \cite{madono2021gan} & 
 \begin{minipage}{2.5cm}
 \centering
 \includegraphics[width=2.5cm]{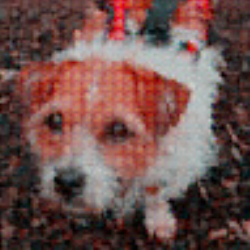}
 \end{minipage} &
 \begin{minipage}{2.5cm}
 \centering
 \includegraphics[width=2.5cm]{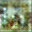}
 \end{minipage} &
 \begin{minipage}{2.5cm}
 \centering
 \includegraphics[width=2.5cm]{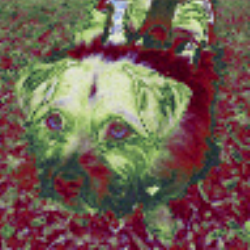}
 \end{minipage} &
 \begin{minipage}{2.5cm}
 \centering
 \includegraphics[width=2.5cm]{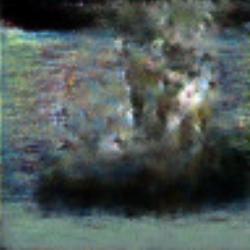}
 \end{minipage} &
 \begin{minipage}{2.5cm}
 \centering
 \includegraphics[width=2.5cm]{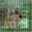}
 \end{minipage} \\
 & 0.864 & 0.009 & 0.109 & 0.058 & 0.123\\[5pt]
 ITN-Attack \cite{itn-attack} & 
 \begin{minipage}{2.5cm}
 \centering
 \includegraphics[width=2.5cm]{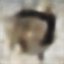}
 \end{minipage} &
 \begin{minipage}{2.5cm}
 \centering
 \includegraphics[width=2.5cm]{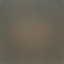}
 \end{minipage} &
 \begin{minipage}{2.5cm}
 \centering
 \includegraphics[width=2.5cm]{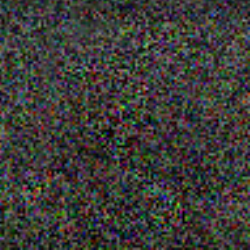}
 \end{minipage} &
 \begin{minipage}{2.5cm}
 \centering
 \includegraphics[width=2.5cm]{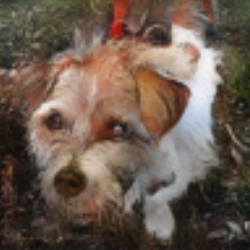}
 \end{minipage} &
 \begin{minipage}{2.5cm}
 \centering
 \includegraphics[width=2.5cm]{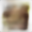}
 \end{minipage} \\
 & 0.281 & 0.071 & 0.017 & 0.664 & 0.167
 \end{tabular}
 \caption{\both{Images restored with three attack methods~\cite{ito2021access}}. Structural similarity index measure (SSIM) values are given under images.}
 \label{fig:sec-eval}
\end{figure*}

Three state-of-the-art attacks, the feature reconstruction attack (FR-Attack)~\cite{fr-attack}, the GAN-based attack (GAN-Attack)~\cite{madono2021gan}, and the inverse transformation network attack (ITN-Attack)~\cite{ito2021image,itn-attack}, were applied to encrypted images for cryptanalysis. As shown in Fig.~\ref{fig:sec-eval}, some visual information on the plain image was restored by using attack methods except network-based transformation~\cite{ito2021image}, where the attacker was assumed to know the encryption algorithms.

\both{We used previously reported results in~\cite{ito2021access}.} Table~\ref{tab:comparison-le} provides a comparison among learnable image transformation methods summarized from a variety of viewpoints. The network-based methods~\cite{sirichotedumrong2020gan,ito2021image} do not require any key, and the block-wise method~\cite{2018-ICCETW-Tanaka} and EtC method~\cite{chuman2019ieeetifs} use a common key, while the extended Tanaka method (E-Tanaka)~\cite{madono2020block} and the pixel-based method~\cite{2019-Access-Warit} utilize a different key. Next, only E-Tanaka~\cite{madono2020block} and EtC~\cite{chuman2019ieeetifs} utilize block shuffling, which is an important step for enhancing robustness against attacks. Another distinct point among encryption methods is the use of an adaptation layer for reducing the influence of encryption in the two block-wise methods~\cite{2018-ICCETW-Tanaka,madono2020block}. In terms of accuracy, the U-Net-based method~\cite{ito2021image} can maintain the classification accuracy that using plain images achieves, but images encrypted by this method cannot be used for training a model. For robustness against attacks, the methods in~\cite{madono2020block,chuman2019ieeetifs} and the U-Net-based approach~\cite{ito2021image} are robust enough against all attacks as shown in Fig.~\ref{fig:sec-eval}.

As summarized above, each encryption method still has some weak points. In particular, encryption methods have to be robust enough against various attacks, and moreover, they should have no performance degradation compared with the use of plain images. In addition, conventional methods focus on image classification, so other privacy-preserving applications such as object detection and semantic segmentation should be discussed.

\begin{table*}
 \centering
 \caption{Comparison of various image transformation methods for privacy-preserving DNNs. $\bigcirc$, $\bigtriangleup$, and $\bigtimes$ denote high, medium, and low, respectively.\label{tab:comparison-le}}
 \begin{tabular}{cccccccc}
 \toprule
 Criteria & Tanaka &
 E-Tanaka &
 Pixel-based & GAN-based & U-Net-based &
 EtC \\
 &
 \cite{2018-ICCETW-Tanaka} &
 \cite{madono2020block} &
 \cite{2019-Access-Warit} &
 \cite{sirichotedumrong2020gan} &
 \cite{ito2021image} &
 \cite{chuman2019ieeetifs}\\
 \midrule
 Block/Pixel key & Common & Different & Different & -- & -- & Common\\[0.5em]
 Block shuffling & -- & \checkmark & -- & -- & -- & \checkmark\\[0.5em]
 Adaptation layer & \checkmark & \checkmark & -- & -- & -- & --\\[0.5em]
 Accuracy & $\bigtriangleup$ & $\bigtriangleup$ & $\bigtriangleup$ & $\bigcirc$ & $\bigcirc$ & $\bigtimes$\\[0.5em]
 Robustness & $\bigtriangleup$ & $\bigcirc$ & $\bigtriangleup$ & $\bigtriangleup$ & $\bigcirc$ & $\bigcirc$\\[0.5em]
 Training/Inference & Both & Both & Both & Both & Inference & Both\\
 \bottomrule
 \end{tabular}
 
\end{table*}

\section{Image transformation for adversarially robust defense}
\label{sec:defense}
Image transformation with a secret key enables us to embed unique features controlled with the key into images. Various applications that do not aim to protect visual information have been inspired by this property. One application is to defend against adversarial examples~\cite{2020-Arxiv-Maung}. 

Intentionally perturbed data points known as adversarial examples are imperceptible to humans, but they cause DNNs to make erroneous predictions with high confidence~\cite{2014-ICLR-Szegedy,Biggio13}. 
To combat adversarial examples, image transformation with a secret key was proposed as a defense~\cite{maung2020icip,2020-Arxiv-Maung,maung2020ensemble}.
The main idea of this defense is to embed a secret key into the model structure with minimal impact on model performance. Assuming the key stays secret, an attacker will not obtain any useful information on the model, which will render adversarial attacks ineffective.

\subsection{Overview of adversarial defense}
\begin{figure}[!t]
\centering\includegraphics[width=\linewidth]{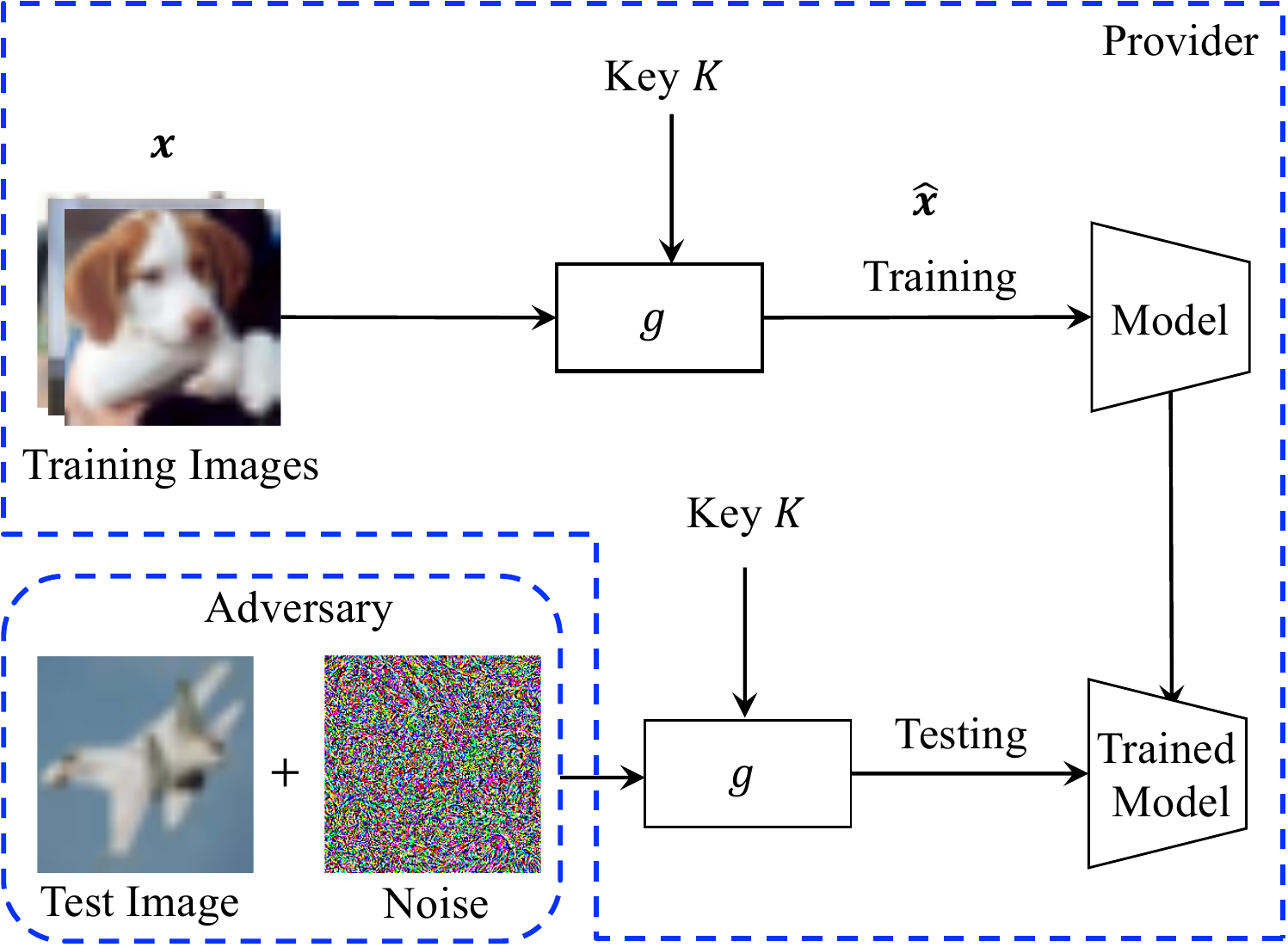}
\caption{Overview of image classification with adversarial defense.\label{fig:def-system}}
\end{figure}

To achieve a defensive ability, a model is trained and tested with transformed images with a secret key. Figure~\ref{fig:def-system} shows an overview of image classification where training and test images are transformed with a secret key. The block-wise transformation used for this application is defined as a function $g$ that takes input $\boldsymbol{x} \in {[0, 1]}^{c \times h \times w}$ for a $c$-channel image of height $h$ and width $w$ and key $K$, and it produces a transformed image $\hat{\boldsymbol{x}}$ (i.e., $g(\boldsymbol{x}, K) = \hat{\boldsymbol{x}}$).

\RB{The goal of an adversarial defense is to have a high classification accuracy for both plain images and adversarial examples. To achieve this goal, the defense proposed in~\cite{maung2020icip,2020-Arxiv-Maung,maung2020ensemble} utilizes an image transformation with a secret key as below.}

\subsection{Transformation procedure}
\label{subsec:block-wise}
\begin{figure*}
\centering
\includegraphics[width=\linewidth]{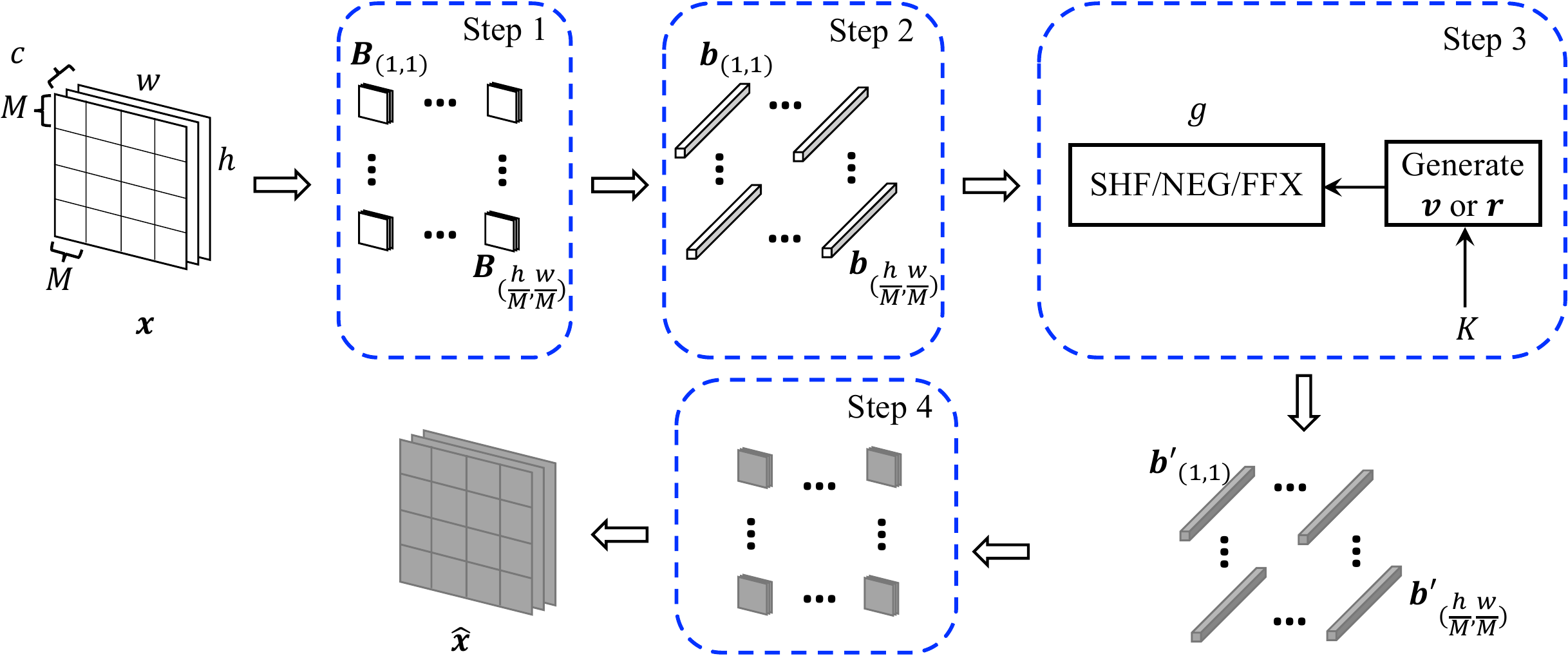} 
\caption{Procedure of block-wise transformation with secret key, $g(\boldsymbol{x}, K, M)$, which takes image $\boldsymbol{x}$, key $K$, and block size $M$ and outputs transformed image $\hat{\boldsymbol{x}}$.\label{fig:blocktrans}}
\end{figure*}

A novel block-wise transformation with a secret key was proposed for adversarial defense as in~\cite{2020-Arxiv-Maung}, where it has three variations: pixel shuffling (SHF), negative/positive transformation (NEG), and format-preserving Feistel-based encryption (FFX)~\cite{2010-NIST-Bellare}. Examples of transformed images for adversarial defense are shown in Fig.~\ref{fig:vis-diff}. The detailed procedure of the block-wise transformation is given as follows (see Fig.~\ref{fig:blocktrans}):

\begin{figure}
\centering
\subfloat[]{\includegraphics[width=0.25\linewidth]{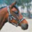}%
\label{fig:horse}}
\hfil
\subfloat[]{\includegraphics[width=0.25\linewidth]{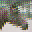}%
\label{fig:horse-s}}
\hfil
\subfloat[]{\includegraphics[width=0.25\linewidth]{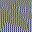}%
\label{fig:horse-i}}
\hfil
\subfloat[]{\includegraphics[width=0.25\linewidth]{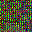}%
\label{fig:horse-e}}
\caption{Example of images generated by three block-wise transformations with $M = 2$. (a) Original image. (b) Pixel shuffling. (c) Negative/positive transformation. (d) format-preserving Feistel-based encryption.\label{fig:vis-diff}}
\end{figure}

\begin{enumerate}
\item Divide $\boldsymbol{x}$ into blocks with a size of $M$ such that $\{\boldsymbol{B}_{(1,1)}, \ldots, \boldsymbol{B}_{(\frac{h}{M}, \frac{w}{M})}\}$.
\item Flatten each block tensor $\boldsymbol{B}_{(i, j)}$ into a vector $\boldsymbol{b}_{(i,j)} = (b_{(i,j)}(1), \ldots, b_{(i,j)}(c \times M \times M))$.

\item Permutate $\boldsymbol{b}_{(i, j)}$ in accordance with the following steps for each transformation.
(For SHF),
\begin{itemize}
\item Generate a random permutation vector $\boldsymbol{v}$ with key $K$, such that 
\\$\boldsymbol{v} = (v_1, \dots, v_k, \dots, v_{k'}, \dots, v_{c \times M \times M})$, where $v_k \neq v_{k'}$ if $k \neq k'$, and $1 \le v_n \le c \times M \times M $.
\item Permutate every vector $\boldsymbol{b}_{(i, j)}$ with $\boldsymbol{v}$ as
 \begin{equation}
 b'_{(i, j)}(k) = b_{(i, j)}(v_k),
 \end{equation}
to obtain a shuffled vector, $\boldsymbol{b}'_{(i, j)} =\\ (b'_{(i,j)}(1), \dots, b'_{(i,j)}(c \times M \times M))$.
\end{itemize}

 (For NEG),
\begin{itemize}
\item Generate a random binary vector $\boldsymbol{r}$ with key $K$, such that 
\\$\boldsymbol{r} = (r_1, \dots, r_k, \dots, r_{c \times M \times M})$, where $r_k \in \{0, 1\}$.
To keep the transformation consistent, $\boldsymbol{r}$ is distributed with \SI{50}{\percent} of ``0''s and \SI{50}{\percent} of ``1''s.
\item Perform negative/positive transformation on every vector $\boldsymbol{b}_{(i, j)}$ with $\boldsymbol{r}$ as
 \begin{equation}
 b'_{(i, j)}(k) = \left\{
 \begin{array}{ll}
 b_{(i, j)}(k) & (r_k = 0)\\
 b_{(i, j)}(k) \oplus (2^L - 1) & (r_k = 1),
 \end{array}
 \right.
 \end{equation}
where $\oplus$ is an exclusive-or (XOR) operation, $L$ is the number of bits used in $b_{(i, j)}(k)$, and $L = 8$ to obtain a transformed vector, $\boldsymbol{b}'_{(i, j)}$.
\end{itemize}

(For FFX),
\begin{itemize}
\item Generate a random binary vector $\boldsymbol{r}$ with key $K$, such that \\$\boldsymbol{r} = (r_1, \dots, r_k, \dots, r_{c \times M \times M})$, where $r_k \in \{0, 1\}$.
To keep the transformation consistent, $\boldsymbol{r}$ is distributed with \SI{50}{\percent} of ``0''s and \SI{50}{\percent} of ``1''s.
\item Convert every pixel value to be at $255$ scale with 8 bits (i.e., multiply $\boldsymbol{b}_{(i, j)}$ by $255$).
\item Perform encryption (FFX) on pixel values in $\boldsymbol{b}_{(i, j)}$ on the basis of $\boldsymbol{r}$ as 
 \begin{equation}
 b'_{(i, j)}(k) = \left\{
 \begin{array}{ll}
 b_{(i, j)}(k) & (r_k = 0)\\
 \text{Enc}(b_{(i, j)}(k)) & (r_k = 1),
 \end{array}
 \right.
 \end{equation}
where $\text{Enc}(\cdot)$ is format-preserving Feistel-based encryption (FFX)~\cite{2010-NIST-Bellare} configured with an arbitrary password and a length of $3$ digits to cover the whole range of pixel values from $0$ to $255$, to obtain a transformed vector, $\boldsymbol{b}'_{(i, j)}$.
\item Convert every pixel value back to $[0, 1]$ scale (i.e., divide $\boldsymbol{b}'_{(i, j)}$ by the maximum value of $\boldsymbol{b}'_{(i, j)}$).
\end{itemize}

\item Integrate the transformed vectors to form a transformed image tensor $\hat{\boldsymbol{x}}$.
\end{enumerate}

\subsection{Robustness against attacks}
Experiments were carried out using projected gradient descent (PGD) under the $\ell_\infty$-norm~\cite{2018-ICLR-Madry}, Carlini and Wagner's attack (CW) under the $\ell_2$-norm~\cite{2017-SP-Carlini}, and elastic net attack (EAD) under the $\ell_1$-norm~\cite{Chen-AAAI-2018} for the CIFAR-10 dataset~\cite{2009-Report-Krizhevsky}.

\begin{figure}[!t]
\centering\includegraphics[width=\linewidth]{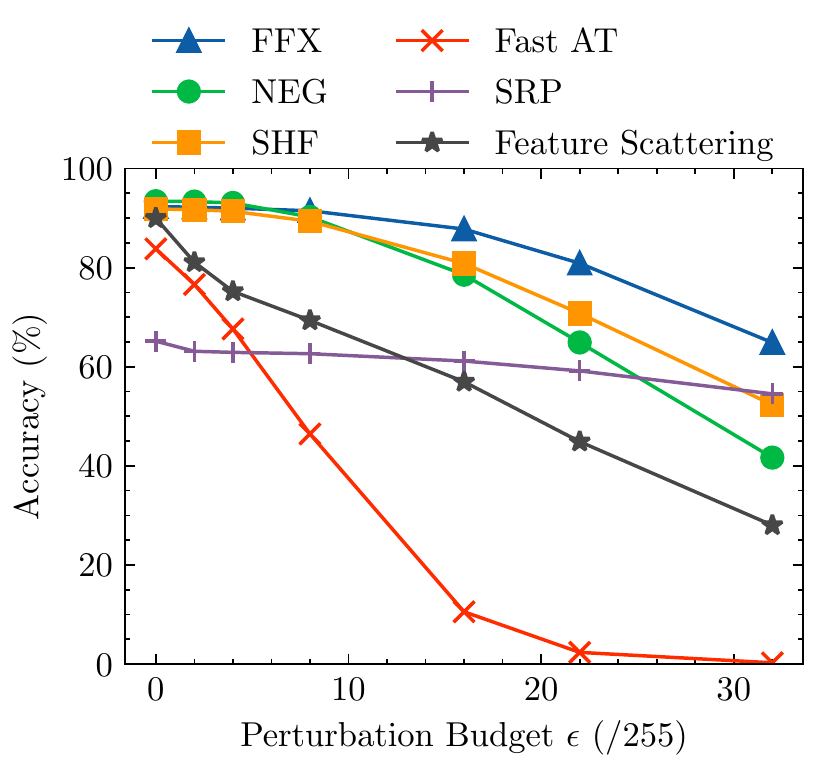}
\caption{\both{Comparison with state-of-the-art defenses in terms of accuracy under PGD attack for CIFAR-10 dataset~\cite{2020-Arxiv-Maung}}. Accuracy was calculated over 10,000 images.\label{fig:def-compare}}
\end{figure}

\both{We used pre-trained models from~\cite{2020-Arxiv-Maung} and reproduced results in} Table~\ref{tab:def-results} that summarizes classification accuracy and attack success rate for models trained by using the three transformations, denoted as SHF, NEG, and FFX, with a block size of 4. From the table, all defense models achieved more than \SI{90}{\percent} accuracy and a low attack success rate (less than \SI{5}{\percent}). Compared with fast adversarial training (Fast AT)~\cite{2020-ICLR-Wong}, the feature-scattering approach (FS)~\cite{2019-NIPS-Zhang}, and standard random permutation (SRP)~\cite{2018-ECCV-Taran} under the PGD attack with various noise distances, the defense models with image transformation achieved a high accuracy as shown in Fig.~\ref{fig:def-compare}.

\sisetup{table-parse-only,detect-weight=true,detect-inline-weight=text,round-mode=places,round-precision=2}
\begin{table}
 \caption{Accuracy and attack success rate of models with adversarial defense under three attacks for CIFAR-10 dataset ($M = 4$)\label{tab:def-results}}
 \centering
 \resizebox{\columnwidth}{!}{%

 \begin{tabular}{lSSSS}
 \toprule
 \multirow{2}{*}{Model} & {Accuracy (\SI{}{\percent})} & \multicolumn{3}{c}{Attack Success Rate (\SI{}{\percent})}\\
 & & {PGD ($\ell_\infty$)} & {CW ($\ell_2$)} & {EAD $\ell_1$)}\\
 \midrule
 SHF & 91.84 & 3.82 & 0.0 & 0.0\\
 NEG & 93.41 & 3.18 & 0.0 & 0.0\\
 FFX & 92.30 & 4.37 & 0.28 & 0.0\\
 \bottomrule
 \end{tabular}
 }
\end{table}

\section{Model protection with image transformation}
\label{seq7}
Training a successful DNN model is not a trivial task and requires three ingredients: a huge amount of data, GPU-accelerated computing resources, and efficient algorithms. For example, the dataset for the ImageNet Large Scale Visual Recognition Challenge (ILSVRC 2012) contains about 1.28 million images, and training on such a dataset takes days and weeks even on GPU-accelerated machines. Considering the expenses necessary to train a DNN model, a trained model should be regarded as a kind of intellectual property (IP).

In this section, we summarize model protection methods. There are two aspects of model protection: model access control and model watermarking. The former focuses on protecting the functionality of DNN models from unauthorized access, and the latter addresses identifying the ownership of models.

\subsection{Model access control}
Chen and Wu first proposed a model access control method that utilizes secret perturbation in such a way that the secret perturbation is crucial to the model's decision~\cite{chen2018protect}. However, this method~\cite{chen2018protect} requires an additional perturbation network, and the parameters of the perturbation network have to be kept secret. In addition, the classification accuracy of the method~\cite{chen2018protect} slightly drops compared with non-protected models under the same training settings. 
\maung{Another method, DeepAttest proposed a hardware-level IP protection to prevent from illegitimate execution of models by using a Trusted Execution Environment (TEE)~\cite{chen2019deepattest}.}
In contrast, a model access control method with image transformation not only does not need any extra network, but it also maintains a high classification accuracy~\cite{maung2021protection}. The model access control method with image transformation is described as follows.

\begin{figure}
\centering
\includegraphics[width=\linewidth]{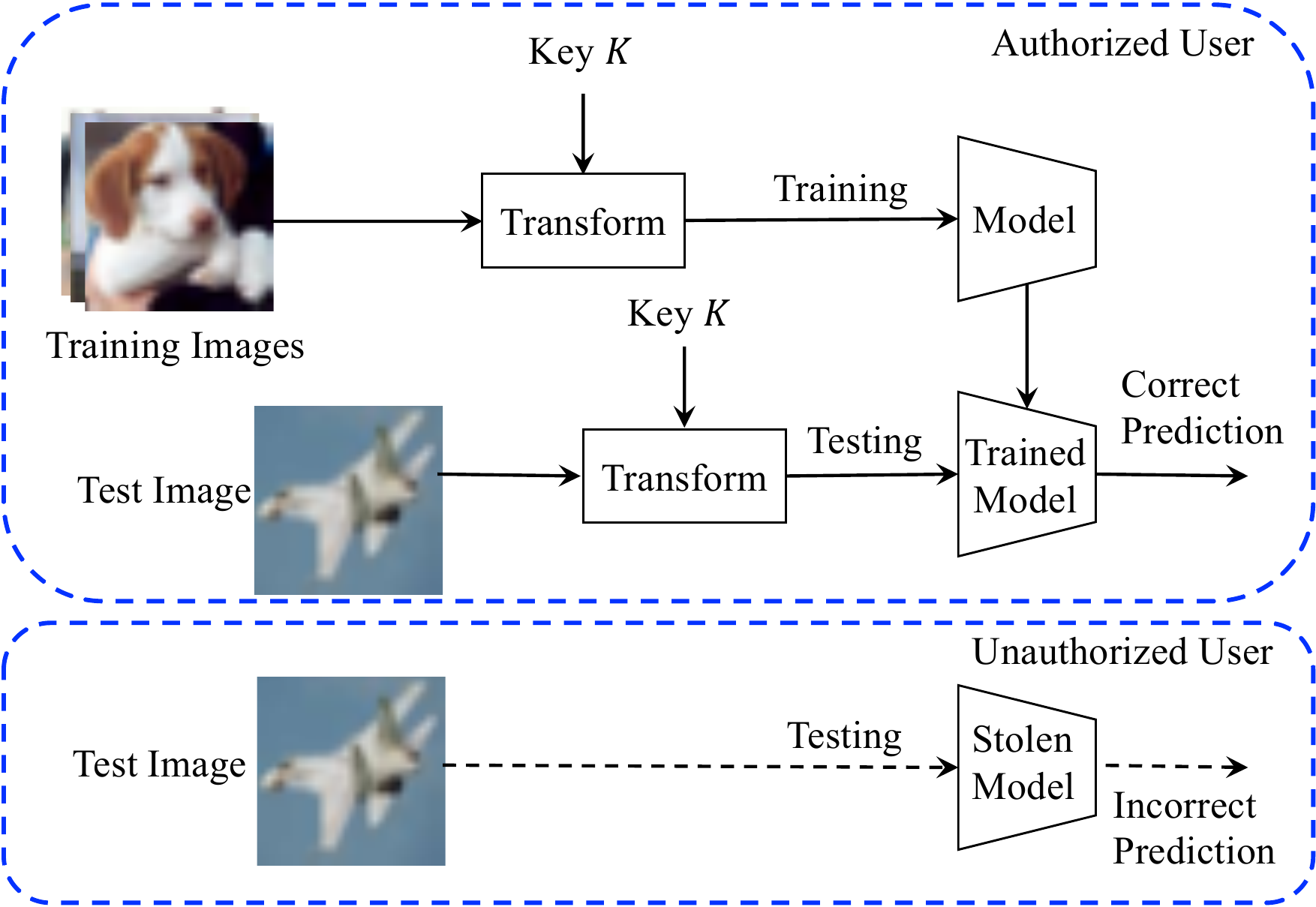} 
\caption{Framework of model access control with image transformation.\label{fig:access}}
\end{figure}

\subsection{Model access control with image transformation}
The block-wise transformations for adversarial defense in Section~\ref{sec:defense} can also be applied to model access control~\cite{maung2021protection}. The framework of model access control with image transformation is shown in Fig.~\ref{fig:access}. From the figure, a model is trained by using images transformed with key $K$. The trained protected model provides correct predictions to authorized users who know the encryption algorithm and correct key $K$. In contrast, unauthorized users cannot use the model to full capacity even when the model is stolen without $K$.

\both{We used pre-trained models with a block size of 4 ($M = 4$) from~\cite{maung2021protection} and reproduced results in} Table~\ref{tab:access-results} that summarizes the results under three conditions: with correct key $K$, with incorrect key $K'$, and without applying transformation to test images (plain) on the CIFAR-10 dataset. The classification accuracy for incorrect key $K'$ was averaged over 100 random keys. The protected models are named after the shorthand for the type of transformation; the model trained by using images transformed by pixel shuffling is denoted as SHF, that by negative/positive transformation as NEG, and that by Feistel-based format preserving encryption as FFX in Table~\ref{tab:access-results}. Among the three models, NEG achieved the best access control performance (i.e., providing a high accuracy for correct key $K$ and a low accuracy for incorrect key $K'$ and plain images).

Moreover, a recent study shows that pixel shuffling as image transformation with a secret key can be applied to one or more feature maps of a network~\cite{maung2021feature,ito2021access}. This approach with feature maps~\cite{maung2021feature} has achieved superior performance compared with using image transformation.

\sisetup{table-parse-only,detect-weight=true,detect-inline-weight=text,round-mode=places,round-precision=2}
\begin{table}
\centering
\caption{Classification accuracy (\SI{}{\percent}) of access control method by image transformation under three conditions ($M = 4$)\label{tab:access-results}}
\begin{tabular}{lSSS}

 \toprule
 {Model} & {Correct ($K$)} & {Incorrect ($K'$)} & {Plain}\\
 \midrule
 {SHF} & 92.58 & 20.15 & 27.77\\
 {NEG} & 93.41 & 12.50 & 12.17\\
 {FFX} & 92.29 & 18.45 & 37.06\\
 \bottomrule
\end{tabular}
\end{table}

\subsection{Model watermarking}
Model watermarking in general aims to identify the ownership of a model only when the model is in question. The functionality of the model is not protected regardless of the ownership. There are two approaches for model watermarking: white-box and black-box. In the white-box approach, a watermark is embedded in model weights by using an embedding regularizer during training. Therefore, access to the model weights is required to extract the watermark embedded in the model as in~\cite{2017-ICMR-Uchida, 2018-Arxiv-Chen, 2018-Arxiv-Rouhani, 2019-NIPS-Fan}. In contrast, in the black-box approach, an inspector observes the input and output of a model in doubt to verify the ownership as in~\cite{2018-USENIX-Yossi,2018-ACCCS-Zhang,2019-NIPS-Fan,2019-MIPR-Sakazawa,2020-NCA-Le,maung2021piracy}. Thus, access to the model weights is not required to verify ownership in black-box approaches.

However, most conventional model watermarking methods are vulnerable to piracy attacks~\cite{wang2019attacks,li2019piracy,2019-NIPS-Fan}. To defend against such attacks, image transformation with a key has been adopted for model watermarking applications~\cite{maung2021piracy} as below.

\subsection{Model watermarking with image transformation}
\begin{figure}
\centering
\includegraphics[width=\linewidth]{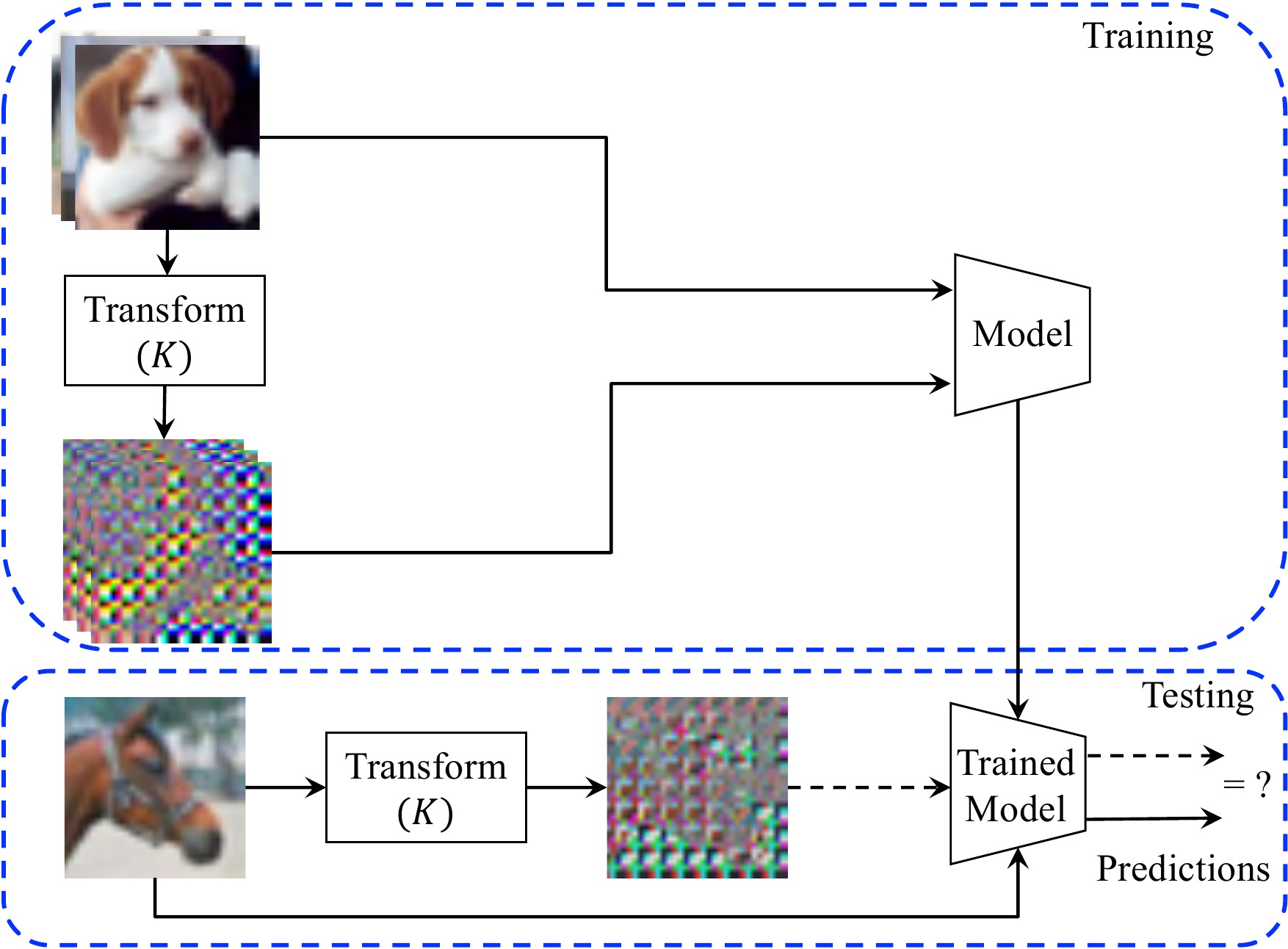} 
\caption{Framework of model watermarking framework with image transformation.\label{fig:watermark}}
\end{figure}

A framework for image classification with model watermarking~\cite{maung2021piracy} is depicted in Fig.~\ref{fig:watermark}. To embed a watermark into a model, the model is trained with both clean images and images transformed with key $K$. Such trained models learn to classify both plain images and transformed ones. The embedded watermark is detected by matching the predictions of the plain and transformed images. This property is exploited to verify the ownership of models. Specifically, watermark detection $\tau$ is defined as
\begin{equation}
 \tau = \frac{1}{N}\sum_{i=1}^{N} \mathbbm{1} (C(f(\boldsymbol{x}_i)) = C(f(\hat{\boldsymbol{x}}_i))), \label{eq:tau}
\end{equation}
where $N$ is the number of test images, $C(f(\cdot))$ is the argmax operation, $f$ is an image classifier, $\boldsymbol{x}_i$ is a test image, $\hat{\boldsymbol{x}_i}$ is a transformed test image, and $\mathbbm{1}(\text{condition})$ is a value of one if the condition is satisfied, otherwise a value of zero.
\maung{To verify the ownership, a user-defined threshold is required.
The threshold should be lower than watermark detection $\tau$.
For example, if the threshold is 80 and $\tau$ is 90, ownership verification is successful because $\tau$ is higher than the threshold.
}

In an experiment, the negative/positive transformation (NEG) in Section~\ref{sec:defense} was applied to the CIFAR-10 dataset.
\both{The results for models with block sizes $M = 2$ and $4$ are taken from~\cite{maung2021piracy} and summarized in Table~\ref{tab:watermark-results}}.
The models were tested with plain images and images transformed by correct key $K$ and incorrect key $K'$, and the corresponding watermark detection values, $\tau$ and $\tau'$, were calculated. When given $K$, both accuracy and watermark detection values were high. In contrast, $\tau'$ dropped significantly under the use of $K'$.

\begin{figure}[t]
\centering
\includegraphics[width=\linewidth]{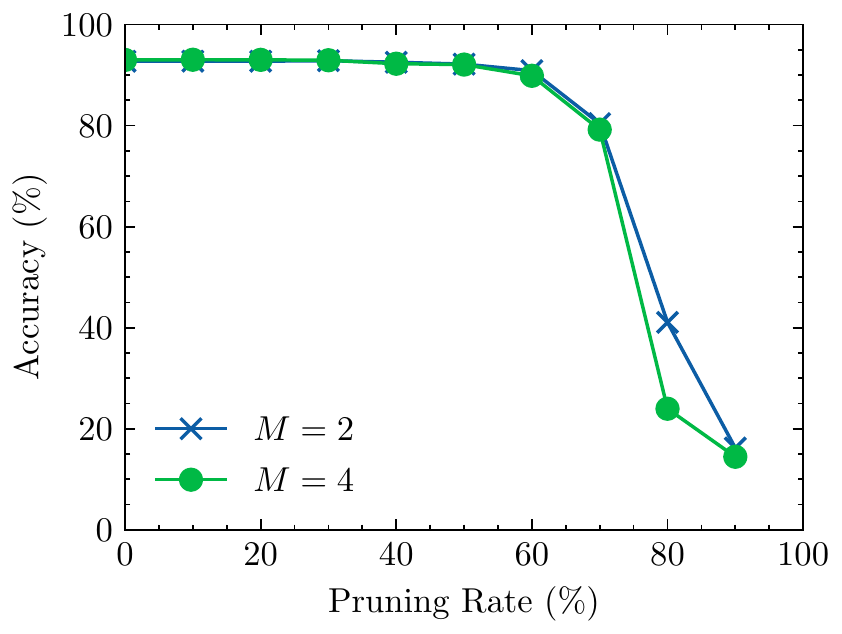} 
\caption{\both{Classification accuracy under pruning attack~\cite{maung2021piracy}}.\label{fig:prune-acc}}
\end{figure}

\begin{figure}[t]
\centering
\includegraphics[width=\linewidth]{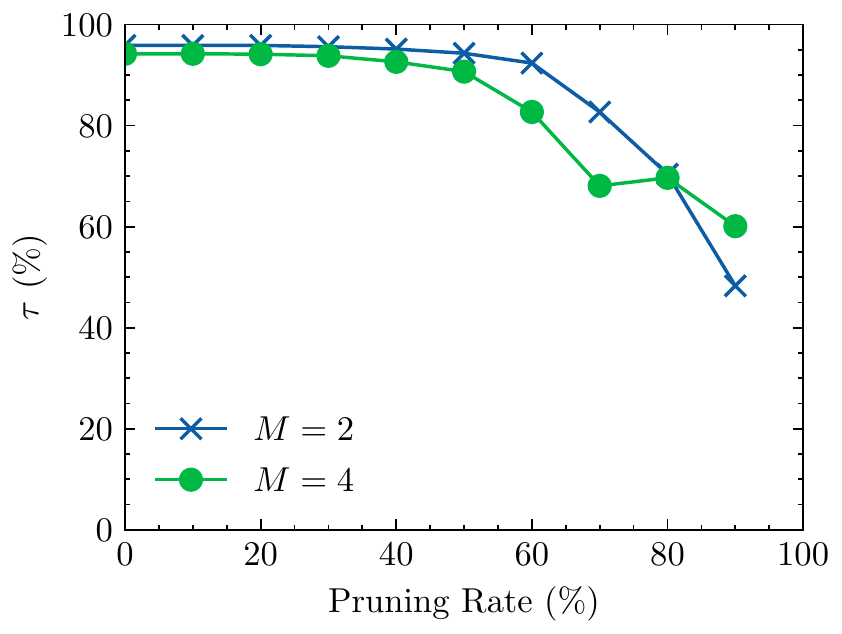} 
\caption{\both{Watermark detection $\tau$ under pruning attack~\cite{maung2021piracy}}.\label{fig:prune-tau}}
\end{figure}

\maung{A watermarked model is pirated if the original watermark is removed or a new verifiable watermark is injected while maintaining a model's accuracy.
Piracy attacks require to modify model weights.
Therefore, Pruning, in which weight values that had the smallest absolute values were zeroed out as in~\cite{2017-ICMR-Uchida}, can be used as a possible attack.}
Figures~\ref{fig:prune-acc} and~\ref{fig:prune-tau} show that watermark detection was resistant up to \SI{60}{\percent} of parameter pruning and was directly dependent on the model accuracy~\cite{maung2021piracy}. Therefore, the results showed that attacking the watermark deteriorated the model accuracy.
\maung{From the results, we can imply that a watermarked model with image transformation cannot be pirated without loosing some accuracy.
}

\sisetup{table-parse-only,detect-weight=true,detect-inline-weight=text,round-mode=places,round-precision=2}
\begin{table}
  \caption{\both{Classification accuracy (\SI{}{\percent}) and watermark detection (\SI{}{\percent}) of protected models with $M = 2$ and $4$~\cite{maung2021piracy}}. Values were averaged over testing whole test set (10,000 images).\label{tab:watermark-results}}
 \centering
 \resizebox{\columnwidth}{!}{%
 \begin{tabular}{l|S|SS|SS}
 \toprule
 & {Accuracy} & {Accuracy} & {Detection} & {Accuracy} & {Detection}\\
 {Model} & {(plain)} & {($K$)} & {($\tau$)} & {($K'$)} & {($\tau'$)}\\
 \midrule
 {$M = 2$} & 92.74 & 93.43 & 95.87 & 10.53 & 10.260\\
 {$M = 4$} & 92.99 & 92.24 & 94.20 & 15.55 & 15.75\\
 \bottomrule
 \end{tabular}
}
\end{table}

\section{Conclusion}
\label{sec:conclusion}
In this paper, we presented an overview of learnable image transformation with a secret key and its applications.
We focused on two properties: compressibility and learnability, although encrypted images have various properties.
The use of these properties allows us not only to compress encrypted images but also to apply them to machine leaning algorithms.
In addition, by using an image transformation method, unique features controlled with a key can be embedded into images, so adversarially robust defenses and model protection can be achieved.

However, conventional transformation methods still have a number of weak points.
Generally, image encryption methods have to be robust enough against various attacks.
\RB{In addition, from the application perspective, when applying encrypted images to privacy-preserving machine learning, classification performance should be maintained compared with the use of plain images.
The use of image transformation with a secret key for adversarial defense and model protection is still at its infancy.
Therefore, there is a lot rooms for improvement in terms of classification accuracy and robustness against various threats.}
In addition, conventional studies have focused on image classification, so other applications such as object detection and semantic segmentation should be discussed as future work.

\bibliographystyle{IEEEtran}
\bibliography{ref}

\end{document}